%% file: acl_latex.tex
\documentclass[11pt]{article}

\usepackage[final]{acl}

\usepackage{times}
\usepackage{latexsym}
\usepackage{booktabs}
\usepackage{array}
\usepackage[breakable]{tcolorbox}
\usepackage[table]{xcolor} 
\usepackage{enumitem}
\usepackage{pifont}        
\usepackage{kotex}
\usepackage{rotating}
\usepackage{amsmath}  

\usepackage[T1]{fontenc}

\usepackage[utf8]{inputenc}

\usepackage{microtype}

\usepackage{inconsolata}

\usepackage{graphicx}
\usepackage{amsfonts}
\usepackage{amsmath}
\usepackage{amssymb}
\usepackage{amsthm}
\usepackage{subcaption}
\usepackage{multirow}

\usepackage{svg} 
\usepackage{etoc}
\usepackage{titletoc}
\usepackage{tikz}
\usetikzlibrary{positioning, arrows.meta, calc}

\usepackage{fvextra}
\fvset{
  breaklines=true,
  breakanywhere=true,
  breaksymbolleft={},
  breaksymbolright={},
  fontsize=\scriptsize,
  xleftmargin=0pt,
  xrightmargin=0pt
}
\usepackage{adjustbox}

\title{NOLLI: A Difficulty-Calibrated English–Korean Puzzle Benchmark}

\author{
Dasol Choi\textsuperscript{1,6}\thanks{Equal contribution.} \quad
Joonyong Park\textsuperscript{2,6}\footnotemark[1] \quad
Daegon Yu\textsuperscript{3,6} \quad
Soo Yong Kim\textsuperscript{6} \\
\textbf{Youngsook Song}\textsuperscript{4,6}\thanks{Corresponding authors.} \quad
\textbf{Seunghyeok Hong}\textsuperscript{5,6}\footnotemark[2] \\[6pt]
{\fontsize{10.5}{12.6}\selectfont
\textsuperscript{1}AIM Intelligence \quad
\textsuperscript{2}KT Corp. \quad
\textsuperscript{3}Sionic AI \quad
\textsuperscript{4}Lablup Inc. \quad
\textsuperscript{5}Hankuk University of Foreign Studies} \\
{\fontsize{10.5}{12.6}\selectfont\textsuperscript{6}HAERAE LAB} \\[5pt]
\raisebox{-0.2em}{\includegraphics[height=1em]{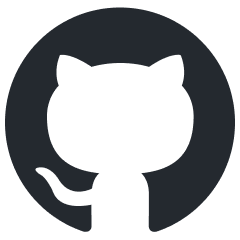}}~\href{https://github.com/HAE-RAE/NOLLI}{GitHub} \quad
\raisebox{-0.2em}{\includegraphics[height=1em]{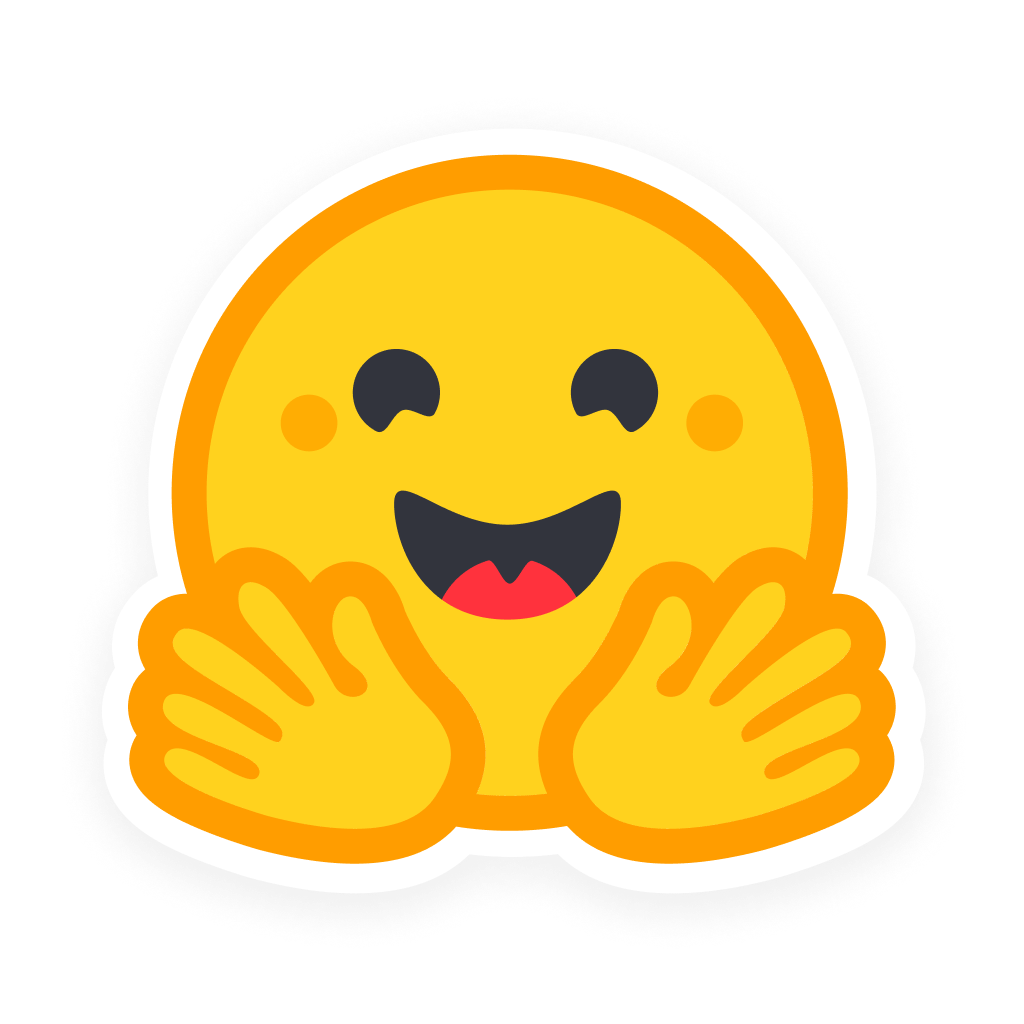}}~\href{https://huggingface.co/datasets/HAERAE-HUB/NOLLI}{HuggingFace} \\[4pt]
{\fontsize{10.1}{10.6}\selectfont
\texttt{dasolchoi@yonsei.ac.kr,\, joonyong.park.dev@gmail.com,\, yssong@lablup.com,\, shongdr@gmail.com}}
}

\usepackage[normalem]{ulem}
\newcommand{\jy}[1]{#1}
\newcommand{\jydel}[1]{}

\begin{document}

\hypersetup{linkcolor=black}
\maketitle

\begin{abstract}
We introduce \textsc{Nolli}, a procedurally generated, difficulty-calibrated English--Korean puzzle benchmark with 15 puzzle types (25 tasks; 7{,}500 items). Each instance is seed-regenerable, has a verified unique solution, and is scored deterministically. We calibrate difficulty behaviorally by tuning generator parameters to target accuracy bands on a reference model. \textsc{Nolli} has three task families: matched direct translations, script adaptations over Hangul \emph{jamo} (sub-syllabic letters), and Korean-only cultural or orthographic tasks. We evaluate 15 frontier, open-weight, and Korean-developed models; gap analyses use the 12 above a 3\% overall-accuracy floor. On matched direct translations, English and Korean accuracy is equivalent within a prespecified $\pm 10$ percentage-point margin. Script adaptations behave differently: Korean Cipher trails English by up to $68.7$ percentage points, while Cryptarithmetic over the same \emph{jamo} shows no systematic Korean penalty, and Jamo Composition strongly correlates with Korean Cipher performance. Korean-only tasks are also heterogeneous: rule-application performance varies across models, while Kinship remains below each non-floor model's Korean direct-task baseline. Finally, structural size does not increase from Easy to Hard in 7 of 15 types, showing that size alone is an unreliable proxy for empirical difficulty.
\end{abstract}

\begin{figure*}[t]
\centering
\includegraphics[width=0.97\textwidth]{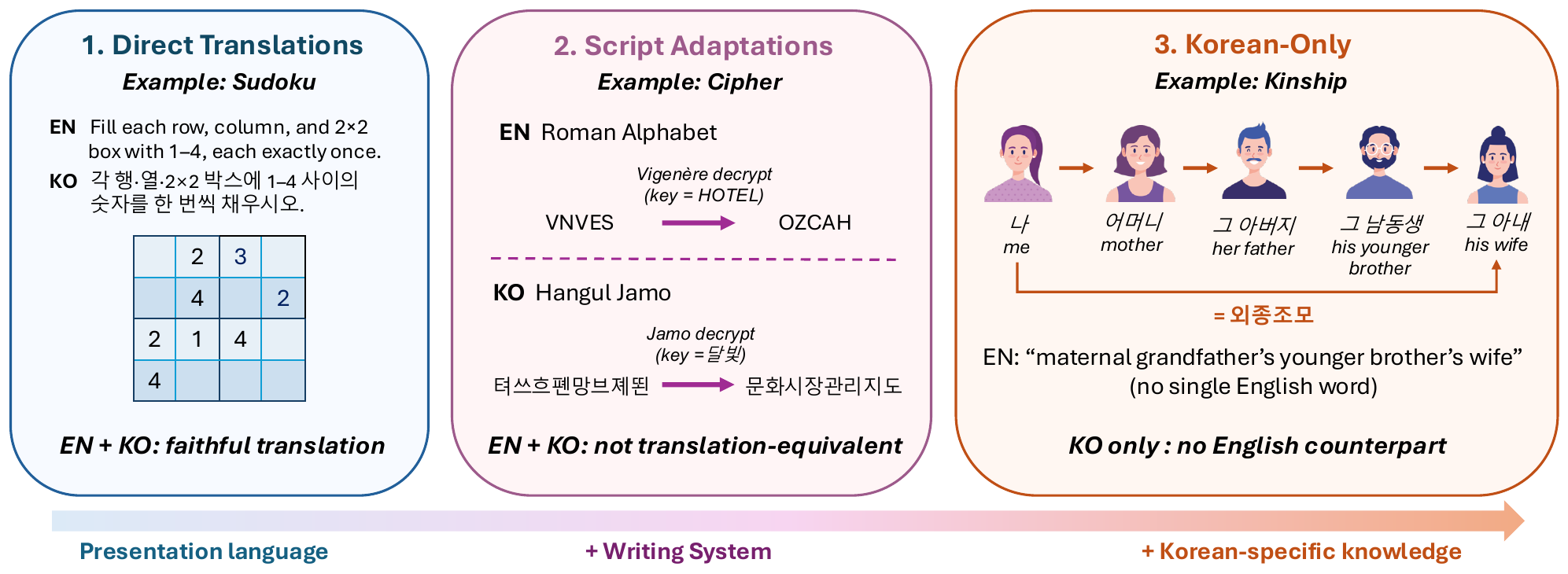}
\caption{The three task families in \textsc{Nolli} (15 types; 25 tasks), with one example each. \textbf{Direct translations}: Sudoku uses the same generator and parameters in English and Korean for matched presentation-language comparison. \textbf{Script adaptations}: Cipher uses Roman letters in English and Hangul \emph{jamo} in Korean and is not translation-equivalent. \textbf{Korean-only}: Kinship maps a four-step relation chain to one Korean term (외종조모) with no single English equivalent. Four Korean-only types are cultural; Jamo Composition is orthographic.}
\label{fig:overview}
\end{figure*}

\section{Introduction}
\label{sec:intro}

Large language models (LLMs) perform increasingly well on reasoning benchmarks, but evaluating this progress remains difficult. Static benchmarks risk contamination and saturation \citep{jacovi2023stop}, while procedurally generated benchmarks often define difficulty using structural properties rather than measured model performance. Multilingual evaluation adds a challenge: some tasks differ only in presentation language, while others also change the script, required knowledge, or reasoning procedure \citep{shi2022language}. A useful multilingual benchmark should therefore support reproducible generation, difficulty control, and a clear distinction between controlled cross-lingual comparisons and language-specific task settings.

We introduce \textsc{Nolli} (Korean for `logic'), a procedurally generated English--Korean benchmark of 15 puzzle types (25 tasks; 7{,}500 items). Every instance is seed-regenerable and deterministically verified to have a unique solution. We tune each generator to target accuracy bands on a fixed reference model, giving the different puzzle types a common behavior-based difficulty scale.

\textsc{Nolli} comprises three task families (Figure~\ref{fig:overview}). Direct translations use matched generator distributions in English and Korean, enabling controlled comparisons of presentation language. Script adaptations operate over Hangul \emph{jamo} (sub-syllabic letters), while Korean-only tasks target cultural or orthographic demands without English counterparts. The latter two families are not controlled translations and are used to examine performance under language-specific task demands.

Across 15 frontier, open-weight, and Korean-developed models, several patterns emerge (gap analyses use the 12 above a 3\% overall-accuracy floor). On direct translations, English--Korean differences range from $-2.7$ to $+5.0$ percentage points and satisfy a prespecified $\pm 10$ percentage-point equivalence margin. Korean Cipher shows gaps of up to $68.7$ percentage points, whereas Cryptarithmetic over the same \emph{jamo} shows no systematic Korean penalty, and Jamo Composition performance is strongly associated with Korean Cipher performance. Korean-only tasks show heterogeneous profiles: rule-application performance varies across models, whereas Kinship remains below each \jy{non-floor} model's Korean direct-task baseline. Finally, the calibrated difficulty ordering transfers broadly across models even though structural size does not increase from Easy to Hard in 7 of 15 puzzle types.

Our main contributions are:
\begin{enumerate}[leftmargin=*, itemsep=2pt]
    \item \textbf{\textsc{Nolli} Benchmark:} A seed-regenerable English--Korean suite of 25 tasks with verified unique solutions and deterministic exact-match evaluation, released with its generators, instances, and evaluators.
    \item \textbf{Behavioral Calibration:} A generator-level protocol that tunes heterogeneous tasks to reference-model target accuracy bands and analyzes how the resulting difficulty ordering transfers across models.
    \item \textbf{Controlled and Language-Specific Evaluation:} A three-family design separating matched presentation-language comparisons from script-adapted and Korean-only task settings.
\end{enumerate}

\section{Related Work}
\label{sec:related}

Static reasoning benchmarks such as GSM8K \citep{cobbe2021gsm8k} and MATH \citep{hendrycks2021math} risk contamination and saturation \citep{jacovi2023stop,zhang2024careful}. Procedurally generated benchmarks address these issues by producing fresh, verifiable instances \citep{sinha2019clutrr,zhu2024dyval,fan2024nphardeval,lin2025zebralogic,wei2025satbench,zhu2025autologi}, but typically control difficulty through generator properties such as problem size or constraint structure. Item Response Theory (IRT) instead estimates difficulty from model responses \citep{vania2021comparing,lalor2024item}, usually after items are generated. Multilingual Reasoning Gym \citep{dobler2026multilingual} extends procedural generation to 94 tasks across 14 languages with parallel instances and adjustable difficulty\jy{, but keeps its cipher task in English script and leaves culture-specific family-relation tasks untranslated}, while MultiZebraLogic \citep{bruun2026multizebralogic} procedurally generates zebra puzzles across nine Germanic languages and studies difficulty through puzzle size and irrelevant red-herring clues. \textsc{Nolli} instead tunes generator configurations to reference-model accuracy bands, making measured model behavior part of the generation process.
\jy{It also targets the regions that parallel translation leaves out, re-implementing script-dependent procedures over Hangul \emph{jamo} and adding Korean-only tasks such as Kinship.}

Multilingual reasoning benchmarks often use translated tasks to compare presentation languages \citep{shi2022language}, while resources emphasize language-specific knowledge or reasoning \citep{son2025kmmlu,kim2024click,son2025hrmcr}. Character-level weaknesses are also well documented \citep{edman2024cute,edman2025execute}, with work examining subword and character representations \citep{xue2022byt5,xu2025enhancing} and Korean tokenization at the \emph{jamo} level \citep{park2020empirical,moon2020jamo}. \textsc{Nolli} complements these lines of work with three task families: matched direct translations for English--Korean comparisons, script adaptations over Hangul \emph{jamo}, and Korean-only cultural or orthographic tasks. This separation keeps controlled cross-lingual comparisons distinct from language-specific evaluation settings.

\section{Benchmark Design}
\label{sec:benchmark}   

\subsection{Terminology and Scale}
\label{sec:terminology}

The benchmark is organized at four levels of granularity.
A \emph{puzzle type} is a problem family with a shared generator and scoring rule (e.g., Sudoku).
A \emph{task} is a type in one language: 10 of the 15 types exist in both English and Korean, and 5 only in Korean. Each task comes in three difficulty \emph{tiers} (Easy, Medium, Hard), and each tier contains 100 generated \emph{instances}.
In short:
\[
\underbrace{(\underbrace{8{\times}2}_{\text{direct}}+\underbrace{2{\times}2}_{\text{script}}+\underbrace{5}_{\text{KO-only}})}_{25~\text{tasks}}
\times\underbrace{3}_{\text{tiers}}\times\underbrace{100}_{\text{inst.}}
=7{,}500.
\]

\subsection{Three Task Families}
\label{sec:crosslingual-design}

\textsc{Nolli} separates controlled presentation-language comparisons from tasks that introduce Korean-specific script, cultural, or orthographic demands. Only direct translations match the underlying task distribution across English and Korean. Script adaptations and Korean-only tasks instead provide targeted evaluation settings in which additional task demands are deliberately introduced.

\paragraph{Direct translations (8 types).}
For these tasks, the underlying puzzle structure and solution are independent of the surface language. The English and Korean conditions use identical generation algorithms and parameter configurations with fixed, author-written language-specific templates; no machine-translation system or LLM is invoked during generation. The conditions are matched in distribution rather than item by item, although three types share latent instances for the paired analysis in Appendix~\ref{app:equivalence}. Appendix~\ref{app:direct-construction} details template construction and validation. Conditional on semantic equivalence between the templates, systematic English--Korean accuracy differences estimate presentation-language effects.

\paragraph{Script adaptations (2 types).}

Cipher and Cryptarithmetic exist in both languages, but the Korean variants are adapted to the compositional structure of Hangul, in which each syllable block decomposes into \emph{jamo}: an initial consonant, a vowel, and optionally a final consonant. Korean Cipher operates over \emph{jamo} indices rather than Roman letters, while Korean Cryptarithmetic maps \emph{jamo} to digits in separate initial, medial, and final tables. Although both operate on the same sub-syllabic units, Cryptarithmetic treats \emph{jamo} as opaque symbols in \jydel{a bijective}\jy{an injective} mapping, whereas Cipher requires explicit decomposition, index arithmetic, and recomposition. The English and Korean variants are deliberately not translation-equivalent and are calibrated independently by language, so their results characterize performance under different script-specific task demands rather than isolate an effect of script alone.

\paragraph{Korean-only tasks (5 types).}

These tasks have no English counterpart. Four draw on Korean cultural systems: Kinship (the Korean kinship-term system), Saju (traditional four-pillars calendar computation), Time (date arithmetic involving the sexagenary day cycle), and Korean Units (arithmetic over traditional measurement units). The fifth, Jamo Composition, tests decomposition and manipulation of Hangul syllable blocks and is orthographic rather than cultural. Because no matched English condition exists, we report each model's performance relative to its accuracy on Korean direct-translation tasks. This within-Korean comparison holds presentation language fixed but remains confounded by task family\jy{, answer space, context length,} and procedural demands\jy{, and for Kinship by lexical knowledge of kinship terms that the prompt does not define}, so it is used descriptively rather than to isolate Korean-specific knowledge.


\jydel{\paragraph{Korean-only tasks (5 types).}}
\jydel{These tasks have no English counterpart. Four draw on Korean cultural systems: Kinship (the Korean kinship-term system), Saju (traditional four-pillars calendar computation), Time (date arithmetic involving the sexagenary day cycle), and Korean Units (arithmetic over traditional measurement units). The fifth, Jamo Composition, instead tests decomposition and manipulation of Hangul syllable blocks and is orthographic rather than cultural. Because a cross-lingual gap is measurable only for the 10 bilingual types, we instead report a within-Korean contrast against each model's accuracy on Korean direct-translation tasks. This holds presentation language fixed but remains confounded by task family and procedural demands, so it does not isolate Korean-specific knowledge.}

\begin{table*}[t]
\centering
\small
\renewcommand{\arraystretch}{0.94}
\begin{tabular}{@{}lp{12.2cm}@{}}
\toprule
\textbf{Type} & \textbf{Task (one-line description)} \\
\midrule
\multicolumn{2}{@{}l}{\emph{Direct translations (8 types; EN{+}KO)}} \\
Array Formula & answer a spreadsheet-style query by filtering, joining, and aggregating relational tables \\
Causal DAG & trace event propagation through a causal graph with AND/OR dependencies \\
Inequality & \jydel{assign the digits 1--9 to variables so that a chain of inequalities holds}\jy{fill an $N{\times}N$ Latin square so that inequality constraints between adjacent cells hold} \\
Minesweeper & identify all mine locations implied by a partially revealed board \\
Number Baseball & infer the secret number from strike/ball feedback on past guesses (Bulls \& Cows) \\
SAT Puzzle & find the truth assignment satisfying a set of natural-language clauses \\
Sudoku & solve the grid and report the values of queried cells \\
Yacht Dice & assign dice rounds to scoring categories for the best total score (Yahtzee-style) \\
\midrule
\multicolumn{2}{@{}l}{\emph{Script adaptations (2 types; EN{+}KO, not translation-equivalent)}} \\
Cipher & decrypt a string under a stack of classical ciphers (EN: Roman letters; KO: jamo) \\
Cryptarithmetic & solve a cryptarithmetic addition and report a queried word's value (EN: letters; KO: jamo) \\
\midrule
\multicolumn{2}{@{}l}{\emph{Korean-only (5 types)}} \\
Kinship & name the Korean kinship term for a person reached via a chain of relations, amid distractors \\
Saju & derive a pillar of the four-pillars (sexagenary) calendar from a date/time or a day pillar \\
Time & resolve relative date statements to a Gregorian date, in part via the sexagenary day cycle \\
Jamo Composition & decompose Hangul syllables into jamo, apply transformations, and recompose \\
Korean Units & convert and sum quantities using a conversion table of traditional units supplied in the prompt \\
\bottomrule
\end{tabular}
\caption{The 15 puzzle types, grouped by cross-lingual category. Appendix~\ref{app:tasks} gives each type's full specification, answer format, per-tier difficulty levers, and a representative instance.}
\label{tab:task-overview}
\end{table*}

\subsection{Task Suite}
\label{sec:tasks}
Table~\ref{tab:task-overview} describes the 15 types; Appendix~\ref{app:tasks} gives each type's full specification, answer format, and a representative instance. The suite spans five reasoning paradigms: constraint satisfaction (SAT Puzzle, Inequality, Minesweeper, Sudoku), algorithmic execution (Cipher, Cryptarithmetic, Array Formula, Jamo Composition, Korean Units), causal and temporal reasoning (Causal DAG, Saju, Time), combinatorial optimization (Yacht Dice), and deductive inference (Number Baseball, Kinship). Answer formats are narrow (an integer, a fixed-format string, a coordinate list, or a variable assignment), so scoring reduces to exact match. To prevent shortcuts, Korean Units supplies a randomized conversion table in the prompt, measuring table-based arithmetic rather than unit recall, while Yacht Dice scores only the globally optimal category assignment, so greedy allocation is insufficient.

\begin{figure}[t]
\centering
\begin{tikzpicture}[
  font=\small,
  pipe/.style={rectangle, rounded corners=4pt, align=center,
               inner sep=5pt, minimum height=0.95cm, line width=0.8pt},
  arr/.style={-{Stealth[length=2.2mm]}, line width=0.9pt, black!75},
]
\node[pipe, draw=cyan!45!blue!65!black, fill=cyan!45!blue!6, text width=3.0cm]
  (gen) {seeded procedural\\generator};
\node[pipe, right=4mm of gen, draw=cyan!45!blue!65!black, fill=cyan!45!blue!13, text width=3.0cm]
  (ver) {unique-solution\\verification};
\node[pipe, below=5mm of ver, draw=cyan!45!blue!65!black, fill=cyan!45!blue!20, text width=3.0cm]
  (ref) {reference model\\{\scriptsize $n{=}100$ per tier}};
\node[pipe, left=4mm of ref, draw=cyan!45!blue!60!black, fill=cyan!45!blue!30, line width=1.1pt, text width=3.0cm]
  (band) {target accuracy bands\\{\scriptsize E 75 / M 50 / H 25 ($\pm$10)}};
\draw[arr] (gen) -- (ver);
\draw[arr] (ver) -- (ref);
\draw[arr] (ref) -- (band);
\draw[arr, dashed, cyan!45!blue!60!black] (band.west) -- ++(-0.35,0) |- (gen.west);
\node[font=\small\itshape, anchor=north, align=center]
  at ($(band.south)!0.5!(ref.south)+(0,-0.15)$) {adjust difficulty levers until in band};
\end{tikzpicture}
\caption{The generation and calibration pipeline. Every instance is procedurally generated with a verified unique solution; each task's three difficulty tiers are calibrated by tuning generator parameters until the reference model's accuracy lands in the target band.}
\label{fig:pipeline}
\end{figure}

\subsection{Procedural Generation and Verification}
\label{sec:generation}

Every instance is produced by a task-specific generator and verified deterministically: backtracking (Sudoku, Inequality, Cryptarithmetic), constraint propagation (Minesweeper), exhaustive filtering (Number Baseball, SAT Puzzle), optimal assignment or simulation (Yacht Dice, Causal DAG, Array Formula), and exact script or calendar computation (Cipher, Jamo Composition, Saju, Time, Kinship, Korean Units). Candidates without a unique answer are rejected, and each released item ships with a solver-generated reference trace.

Each task ships with a deterministic evaluator that extracts the declared final answer, normalizes whitespace, unordered collections, and numeric formatting, canonicalizes structured outputs, and scores by exact match against the gold answer. No LLM judge is used anywhere in the pipeline, so scoring is fully reproducible.

\begin{table*}[t]
\centering
\footnotesize
\renewcommand{\arraystretch}{0.97}
\begin{tabular}{@{}l @{\hspace{7pt}} ccc @{\hspace{9pt}} ccc @{\hspace{9pt}} ccc @{\hspace{9pt}} cc @{\hspace{9pt}} c@{}}
\toprule
\multirow{2}{*}{\textbf{Model}} & \multicolumn{3}{c}{\multirow{2}{*}{\textbf{Direct}}} & \multicolumn{6}{c}{\textbf{Script adaptations}} & \multicolumn{2}{c}{\multirow{2}{*}{\textbf{Korean-only}}} & \multirow{2}{*}{\textbf{Avg}} \\
\cmidrule(lr){5-10}
 & \multicolumn{3}{c}{} & \multicolumn{3}{c}{Cipher} & \multicolumn{3}{c}{Cryptarith.} & \multicolumn{2}{c}{} & \\
\cmidrule(lr){2-4} \cmidrule(lr){5-7} \cmidrule(lr){8-10} \cmidrule(lr){11-12}
 & EN & KO & $\Delta$ & EN & KO & $\Delta$ & EN & KO & $\Delta$ & Cult (4) & Jamo & \\
\midrule
\multicolumn{13}{@{}l}{\emph{Frontier API}} \\
GPT-5.5 & \textbf{83.4} & \textbf{82.3} & $+1.0$ & \textbf{95.3} & \textbf{93.7} & $+1.7$ & \textbf{96.0} & \textbf{97.3} & $-1.3$ & \textbf{81.8} & \textbf{87.3} & \textbf{84.9} \\
Opus 4.8 & 79.6 & 77.4 & $+2.2$ & 93.0 & 69.3 & $+23.7$ & 89.7 & 95.0 & $-5.3$ & 77.2 & 47.0 & 78.3 \\
Gemini 3.1 & 75.8 & 73.7 & $+2.1$ & 84.7 & 63.0 & $+21.7$ & 82.0 & 88.0 & $-6.0$ & 67.6 & 57.3 & 73.6 \\
\addlinespace[1.5pt]
\multicolumn{13}{@{}l}{\emph{International open-weight}} \\
Qwen3.5-9B & 39.8 & 38.8 & $+1.0$ & 27.7 & 5.7 & $+22.0$ & 32.0 & 23.0 & $+9.0$ & 20.5 & 12.3 & 32.5 \\
Qwen3.5-27B & 58.3 & 53.5 & $+4.8$ & 35.3 & 0.0 & $+35.3$ & 39.3 & 6.7 & $+32.7$ & 33.1 & 1.3 & 44.4 \\
Qwen3.5-397B & 68.3 & 69.5 & $-1.2$ & 65.0 & 26.7 & $+38.3$ & 59.7 & 65.0 & $-5.3$ & 49.5 & 46.3 & 62.5 \\
Gemma-4-31B & 56.3 & 55.4 & $+0.9$ & 66.3 & 8.7 & $+57.7$ & 42.0 & 43.0 & $-1.0$ & 35.2 & 13.7 & 48.3 \\
gpt-oss-120b & 56.6 & 55.1 & $+1.5$ & 55.7 & 0.7 & $+55.0$ & 48.3 & 15.3 & $+33.0$ & 42.2 & 8.7 & 47.7 \\
DS-V4-Flash & 49.5 & 44.5 & $+5.0$ & 74.3 & 5.7 & $+68.7$ & 52.7 & 56.0 & $-3.3$ & 46.4 & 17.3 & 45.8 \\
Llama-3.1-8B & 2.3 & 1.2 & $+1.1$ & 0.7 & 0.0 & $+0.7$ & 1.3 & 1.3 & $+0.0$ & 1.5 & 1.0 & 1.5 \\
L4-Maverick & 20.6 & 18.9 & $+1.7$ & 0.7 & 4.3 & $-3.7$ & 5.7 & 1.3 & $+4.3$ & 7.3 & 2.0 & 14.4 \\
\addlinespace[1.5pt]
\multicolumn{13}{@{}l}{\emph{Korean-developed}} \\
EXAONE-4.0 & 30.9 & 30.2 & $+0.7$ & 35.7 & 0.0 & $+35.7$ & 33.3 & 0.0 & $+33.3$ & 17.8 & 0.0 & 25.1 \\
Solar-100B & 14.7 & 17.4 & $-2.7$ & 21.3 & 0.0 & $+21.3$ & 3.3 & 2.3 & $+1.0$ & 15.9 & 3.7 & 14.1 \\
Mi:dm-2.0 & 0.8 & 1.0 & $-0.2$ & 0.0 & 0.0 & $+0.0$ & 0.0 & 0.0 & $+0.0$ & 8.9 & 0.7 & 2.0 \\
EXAONE-3.5 & 0.5 & 0.4 & $+0.1$ & 0.0 & 0.0 & $+0.0$ & 0.0 & 0.0 & $+0.0$ & 3.8 & 0.0 & 0.9 \\
\addlinespace[1.5pt]\midrule
\emph{Reference} & 52.4 & 52.5 & $-0.1$ & 51.0 & 54.7 & $-3.7$ & 48.7 & 55.0 & $-6.3$ & 52.0 & 50.7 & 52.3 \\
\bottomrule
\end{tabular}
\caption{Mean accuracy (\%) by task group, pooled over difficulty tiers. $\Delta=\text{EN}-\text{KO}$ in percentage points, computed from unrounded accuracies (displayed subtraction may differ by $0.1$\,pp); bold marks the best non-reference value in each column. Cult = four Korean cultural-system tasks; Jamo = Jamo Composition; Avg = 25-task macro average. Reference = the calibration model, Gemini 3 Flash; its script-adaptation gaps reflect independent calibration by language. Full per-task and per-tier results appear in Appendix~\ref{app:full-results}.}
\label{tab:main-results}
\end{table*}

\subsection{Difficulty Calibration by Target Accuracy Bands}
\label{sec:calibration}

Procedural benchmarks often define difficulty through structural generator parameters (e.g., ``hard'' = larger grid). We instead define difficulty \emph{behaviorally}: Easy, Medium, and Hard are reference-relative tiers determined by model accuracy, with generator parameters tuned to place each tier in a target band.

\paragraph{Protocol.}
We use Gemini 3 Flash \citep{google2025gemini3flash} at reasoning effort medium as the reference model\jy{, chosen for its balance of reasoning strength and cost across the many evaluations that calibration requires,} and set target accuracy bands of $75\pm10$\% (Easy), $50\pm10$\% (Medium), and $25\pm10$\% (Hard), measured over $n{=}100$ instances per tier. Calibration proceeds iteratively (Figure~\ref{fig:pipeline}): for each task, we select generator parameters as difficulty levers, sweep them, evaluate the reference model, and adjust them until (i) each tier falls inside its band and (ii) adjacent tiers differ by at least 10\jydel{\,pp} \jy{percentage points (pp)}. For direct-translation tasks, calibration runs on the English variant and the resulting configuration is mirrored to Korean, preserving parameter equivalence. Script adaptations are calibrated independently by language because they are not translation-equivalent; Korean-only tasks are calibrated directly in Korean.

\paragraph{Outcome.}
Of the 75 task--tier combinations, 72 fall inside their target bands (Table~\ref{tab:calibration} in Appendix~\ref{app:calibration}). All 51 directly calibrated tiers are in band with adjacent tiers separated by at least 10\,pp. The three misses occur among the 24 mirrored Korean tiers and exceed their upper boundaries by 2--7\,pp (SAT Puzzle Easy 87\%, Yacht Dice Hard 38\%, and Number Baseball Hard 42\%). We retain these mirrored configurations because re-tuning Korean independently would break parameter equivalence. For the independently calibrated script adaptations, average reference-model accuracy differs modestly by language (Cryptarithmetic: $48.7$ EN vs.\ $55.0$ KO; Cipher: $51.0$ vs.\ $54.7$), with both differences favoring Korean.

\paragraph{Reference-model sensitivity.}
To test whether tier construction depends strongly on the chosen reference, we recalibrated four representative conditions (Causal DAG EN, Cipher EN/KO, and Time KO) using Qwen3.5-27B and DeepSeek-V4-Flash. Both alternative references reached the same target bands, and cross-evaluation preserved Easy $>$ Medium $>$ Hard in 7 of 8 configurations \jy{(Table~\ref{tab:ref-sensitivity}, Appendix~\ref{app:calibration})}. The exception was \jy{DeepSeek-calibrated Korean Cipher evaluated by Qwen3.5-27B,} a near-floor regime (11/18/0\%). \jy{Qwen3.5-27B also solved none of the original Korean Cipher items (0/0/0\%), so its Korean Cipher tiers mix in reverse-only items (80/54/27\% of Easy/Medium/Hard) instead of retuning the original configuration.} These results support ordinal transfer across references, while the absolute bands remain reference-relative.

\paragraph{Regeneration.}
Because difficulty is tied to calibrated generator configurations rather than a fixed item set, fresh instances can be sampled from each tier. In a held-out check on six representative task types (eight language-specific tasks; 24 task--tier cells), all eight pairs preserved Easy $>$ Medium $>$ Hard, and 20 of 24 cells (83\%) remained within their target bands; the four misses were at most 7\,pp (Appendix~\ref{app:regeneration}). We therefore treat regeneration as preserving tier ordering and approximate, rather than guaranteed, absolute difficulty. Released configurations are frozen within each benchmark version; future recalibration would define a new version rather than alter existing tiers\jy{, and English--Korean gaps are compared within a version}.

\begin{figure*}[t]
\centering
\setlength{\abovecaptionskip}{6pt}
\includegraphics[width=\textwidth]{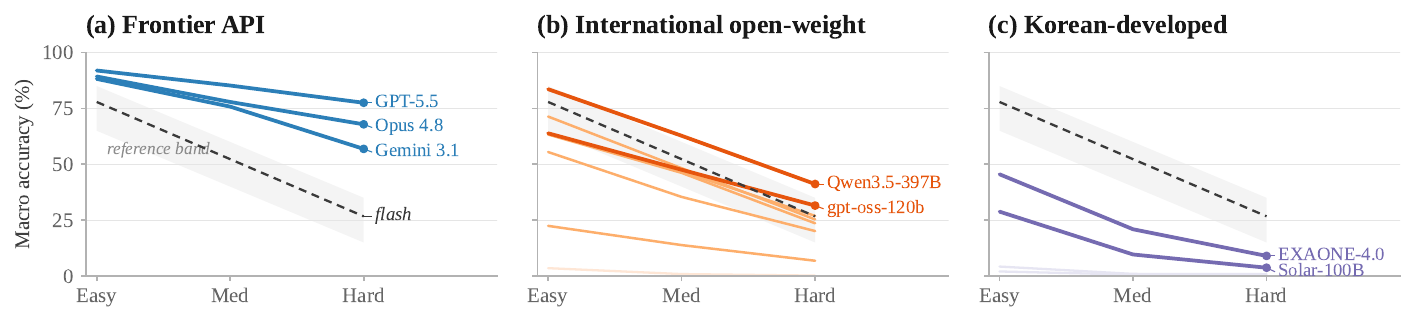}
\caption{Macro accuracy by difficulty tier, faceted by developer group. Shaded ribbon:
reference model's target band (65--85 / 40--60 / 15--35); dashed line: the reference
model (\texttt{gemini-3-flash}), whose tier accuracies fall within the target bands by construction. Saturated lines are
labeled exemplars; faintest lines are floor models ($<$3\%). Though the bands are
anchored to one reference model, the difficulty \emph{ordering} transfers: accuracy
decreases Easy$\to$Hard for 255 of 300 non-floor model--task pairs (85\%; 94\% with
ties).}
\label{fig:tiers}
\end{figure*}

\section{Experimental Setup}
\label{sec:experiments}

\paragraph{Models}

We evaluate 15 models spanning three groups: three frontier API models, eight international open-weight models, and four Korean-developed models (Table~\ref{tab:main-results}). Full model identifiers, references, parameter counts, and inference configurations are provided in Appendix~\ref{app:model-configs}.

\paragraph{Implementation Details}





All models are evaluated with HRET \citep{hret}. Proprietary models use native APIs; open-weight models are served with vLLM or accessed via OpenRouter. We follow recommended sampling configurations, using medium reasoning effort where available and each model's supported reasoning mode otherwise. The output budget is 32{,}768 tokens, reduced to 14{,}336 for the two 32k-context models (Mi:dm-2.0 and EXAONE-3.5). Full inference configurations and exceptions appear in Appendix~\ref{app:model-configs}.

\section{Results and Analysis}
\label{sec:analysis}

\subsection{Overall Performance}
\label{sec:overall-performance}
Table~\ref{tab:main-results} reports accuracy across the cross-lingual task spectrum over the full 7,500-item benchmark.
Frontier API models achieve the highest macro accuracies, led by GPT-5.5 (84.9\%), Opus 4.8 (78.3\%), and Gemini 3.1 (73.6\%).
Among international open-weight models, accuracy tracks capacity within the Qwen family (32.5\% at 9B to 62.5\% at 397B), with Qwen3.5-397B reaching near-parity on direct translations (68.3\% EN vs.\ 69.5\% KO).
Mid-sized models (Gemma-4-31B 48.3\%, gpt-oss-120b 47.7\%, DeepSeek-V4-Flash 45.8\%) perform reliably on direct translations but degrade sharply on sub-syllabic tasks.
Korean-developed models display a distinct profile.
EXAONE-4.0 (25.1\%) and Solar-100B (14.1\%) underperform international open-weight models of similar scale (e.g., Qwen3.5-27B at 44.4\%), and EXAONE-3.5 (0.9\%) and Mi:dm-2.0 (2.0\%) fall below 3\% overall, as does Llama-3.1-8B (1.5\%).

Figure~\ref{fig:tiers} shows that the calibrated tier ordering transfers broadly beyond the reference model: among the 12 non-floor models (above $3\%$ overall accuracy), accuracy decreases from Easy to Hard for 255 of 300 model--task pairs (85\%; 94\% including ties).
Absolute levels differ by developer group, but the tier structure transfers.

\subsection{Analysis Across Task Families}
\label{sec:cross-lingual-decomposition}

We analyze Korean performance differences across three regimes: presentation language under matched task distributions, writing-system-intensive processing, and Korean-specific knowledge; the latter two provide diagnostic rather than causal comparisons.

\paragraph{Direct Translations: Equivalence Within the Calibration Margin.}
On direct-translation tasks (8 puzzle types), English and Korean instances are drawn from the same generator under identical parameters, so, conditional on semantic equivalence of the templates, their accuracy difference estimates the effect of presentation language. Across the 12 non-floor models, English--Korean gaps range from $-2.7$ to $+5.0$\,pp, and Holm-corrected TOSTs \jy{(two one-sided tests for equivalence)} support equivalence within the $\pm 10$\,pp calibration margin. The presentation-language effect is thus smaller than the resolution at which the benchmark defines task difficulty, from frontier to Korean-developed models. On the three tasks with item-level correspondence, paired TOST reproduces the equivalence; uncorrected McNemar tests flag nominal differences for four models, all small ($\le 4.8$\,pp) and within the equivalence margin (Appendix~\ref{app:equivalence}).

\paragraph{Script Adaptations: Failures Localize to Syllable Decomposition and Recomposition.}

Cryptarithmetic and Cipher provide a diagnostic contrast in sub-syllabic manipulation. Both operate over Hangul \emph{jamo} but impose different procedural demands: Cryptarithmetic treats \emph{jamo} as opaque symbols in \jydel{a bijective}\jy{an injective} mapping, whereas Cipher requires explicit disassembly, index arithmetic, and recomposition.

Cryptarithmetic shows no systematic Korean penalty: 6 of 12 non-floor models score \emph{higher} in Korean, including all three frontier models. A substantial penalty appears in only three (Qwen3.5-27B, gpt-oss-120b, and EXAONE-4.0; all near $+33$\,pp), consistent with difficulty in treating \emph{jamo} as opaque variables. Cipher collapses instead. Every non-frontier model with nontrivial English accuracy loses at least $21$\,pp, with near-total drops for four: Qwen3.5-27B, EXAONE-4.0, Solar-100B, and gpt-oss-120b score $21.3\text{--}55.7\%$ in English but below $1\%$ in Korean. DeepSeek-V4-Flash shows the largest gap ($74.3\%$ EN vs.\ $5.7\%$ KO, $+68.7$\,pp), despite a $-3.7$\,pp reference-model residual in the opposite direction. Among the 12 non-floor models, Llama-4-Maverick alone scores higher in Korean ($4.3\%$ vs.\ $0.7\%$ EN). Both scores are near floor, and this reversal may partly reflect answer-space asymmetry in the Easy tier: Korean plaintexts follow a fixed four-slot template, whereas English plaintexts are random letter strings (Appendix~\ref{app:tasks}). Even frontier models retain a sizable gap, with only GPT-5.5 closing it ($+1.7$\,pp).

Success on Cryptarithmetic shows that subword tokenization is not an absolute barrier to operating over \emph{jamo}. Across the 12 non-floor models, Jamo Composition accuracy predicts Korean Cipher accuracy ($R^2=0.91$); the relationship remains strong after controlling for Direct-KO accuracy (partial $r=0.87$, $p<.01$). This contrast is consistent with difficulty in multi-step sub-syllabic execution\jydel{, although the correlation does not establish the mechanism}\jy{; an operation-level probe (Section~\ref{sec:cipher-errors}) localizes it to syllable decomposition and recomposition}.

\paragraph{Korean-Only Tasks: Mixed Rule-Application Deficits, Consistent Kinship Deficit.}
The four Korean-only tasks other than Jamo Composition fall into two regimes: Saju, Time, and Korean Units test deterministic rule application (with the Korean Units conversion table supplied in the prompt), whereas Kinship requires mapping relations to Korean kinship terms whose meanings are not provided (26 candidates per item; uniform chance $3.8\%$). We therefore report the within-Korean deficit (Direct-KO accuracy minus task accuracy) separately for the two groups. These contrasts are descriptive: because the tasks differ from the Direct-KO baseline in more than cultural knowledge, they do not identify a cultural-knowledge effect in isolation.

The rule-application deficit varies in sign and does not vary monotonically with overall benchmark performance. It becomes a surplus for four models spanning the performance range (GPT-5.5: $-11.6$\,pp; DeepSeek-V4-Flash: $-10.6$\,pp; Opus 4.8: $-4.4$\,pp; Solar-100B: $-2.4$\,pp), while remaining positive at all three Qwen scales ($+14.5$, $+15.6$, and $+13.3$\,pp for 9B$\to$27B$\to$397B).

The Kinship deficit, by contrast, does not vary in sign. It remains positive in all 12 non-floor models and shows no evidence of narrowing with overall benchmark performance ($r{=}0.16$, $p{=}.62$, $n{=}12$): models improve on Kinship without systematically closing the distance to their own Korean baseline. Within the frontier group, the Kinship ordering inverts the overall one: Gemini 3.1 (69.0\%) $>$ Opus 4.8 (63.3\%) $>$ GPT-5.5 (45.7\%). The extremes have non-overlapping Wilson intervals ($[63.6, 74.0]$ vs.\ $[40.1, 51.3]$; $n{=}300$). GPT-5.5, the suite's highest-scoring model, thus carries a $+36.7$\,pp Kinship deficit, whereas Gemini 3.1, the lowest-scoring frontier model, nearly closes the gap ($+4.7$\,pp).

\subsection{Where Difficulty Comes From}
\label{sec:difficulty-mechanisms}

Reasoning benchmarks often operationalize difficulty through structural size variables, such as grid dimensions, context length, or variable count. Our calibration results show that size alone is insufficient to characterize empirical difficulty. As Table~\ref{tab:difficulty-mechanisms} summarizes, a salient size measure grows from Easy to Hard in 8 of the 15 puzzle types. For 5 other types, difficulty rises without monotonic growth in the corresponding size measure, relying instead on constraint density, information masking, or task-type reweighting. In the remaining 2 types, a salient length measure decreases even as empirical difficulty rises. Taken together, a salient size measure fails to grow from Easy to Hard in 7 of the 15 types, nearly half the suite.

\begin{table}[t]
\centering
\setlength{\tabcolsep}{2.5pt}
\renewcommand{\arraystretch}{0.95}
{\fontsize{8.5}{10.2}\selectfont
\begin{tabular}{@{}l p{\dimexpr\columnwidth-2.2cm\relax}@{}}
\toprule
\textbf{Task} & \textbf{Knob (Easy $\to$ Hard)} \\
\midrule
\multicolumn{2}{@{}l}{\emph{Salient size measure grows (Easy $\to$ Hard)}} \\
SAT Puzzle & Variables $9 \to 14$; clauses $36\text{--}54 \to 115\text{--}170$ \\
Cryptarithm. & Operand length $6 \to 8$ letters (EN), $6 \to 7\text{--}8$ \emph{jamo} (KO) \\
Array Formula & Order rows $35\text{--}50 \to 220\text{--}300$; distractor cols $0 \to 11$ \\
Causal DAG & Events $25\text{--}31 \to 46\text{--}58$; edge density $.62 \to .95$ \\
Jamo Comp. & Syllables $2 \to 3$; final consonant light $\to$ mixed \\
Korean Units & Summed items $17 \to 26$; max coefficient $1{,}249 \to 6{,}999$ \\
Inequality & Grid $5{\times}5 \to 6{\times}6$ ($6{\times}6$ share $0 \to .80$) \\
Number Baseball & Digits $7 \to 8$; revealed digits $5 \to 0$ \\
\midrule
\addlinespace[2.5pt]
\multicolumn{2}{@{}l}{\emph{Size measure fixed or non-monotonic}} \\
Sudoku & Grid fixed at $9{\times}9$; givens $41 \to 33$ \\
Minesweeper & Grid $7{\times}7\text{--}12{\times}12 \to 9{\times}9$; revealed cells $\downarrow$ \\
Yacht Dice & 12 rounds fixed; greedy--optimal gap $0\text{--}8 \to {\geq}30$ \\
Saju & Internal subtype mix (med.\ : hard) $4{:}1 \to 1{:}8$ \\
Time & Instance length fixed; \emph{ganji} fraction $.34 \to .91$ \\
\midrule
\addlinespace[2.5pt]
\multicolumn{2}{@{}l}{\emph{Salient length measure shrinks ($\downarrow$)}} \\
Cipher (EN) & Answer $20\text{--}24 \to 6\text{--}10$ chars; cipher layers $1 \to 2$ \\
Kinship & Relation hops $4\text{--}5 \to 2\text{--}3$; distractor utterances $15 \to 112$ \\
\bottomrule
\end{tabular}}
\caption{Difficulty levers grouped by how a salient size measure changes from Easy to Hard. The first knob in each row defines the grouping; Cipher refers to the independently calibrated English variant. Full configurations appear in Appendix~\ref{app:tasks}.}
\label{tab:difficulty-mechanisms}
\end{table}

\subsubsection{Taxonomy of Escalation Levers}

We identify four operational mechanisms used to construct the difficulty tiers. These mechanisms are not mutually exclusive: some generators vary more than one lever.

\begin{enumerate}[leftmargin=*, itemsep=1pt]
    \item \textbf{Size Scaling:} Expanding the structural scale of a task by increasing variable counts, sequence lengths, or tabular dimensions (e.g., SAT Puzzle, Cryptarithmetic, and Korean Units).
    \item \textbf{Constraint and Information Structure:} Holding the core grid or state horizon fixed while withholding clues or changing the constraint structure. Reducing Sudoku givens from $41$ to $33$ removes constraint information on the same $9{\times}9$ board, while favoring generic Yacht Dice categories widens the gap between greedy and optimal play over a fixed 12-round horizon.
    \item \textbf{Distractor Density:} Injecting irrelevant information into the prompt to test selective attention. In Array Formula, adding up to 11 non-queryable distractor columns requires the model to filter noise before executing aggregations.
    \item \textbf{Algorithmic and Task-Type Depth:} Requiring additional procedural transformations or shifting the task mixture toward unassisted computations (e.g., Saju and Time shifting toward from-scratch sexagenary calendar arithmetic).
\end{enumerate}

\subsubsection{Counterexamples: Input Size Is Not Sufficient}

Several cases show why simple input-size measures are unreliable proxies for LLM reasoning difficulty.

\paragraph{Kinship}
Reasoning-chain length decreases from $4\text{--}5$ hops in Easy to $2\text{--}3$ in Hard, yet reference accuracy falls from 77\% to 31\%. The tier shift instead coincides with increased distractor dialogue ($15 \to 112$ utterances) and a greater prevalence of close-kin relations involving subtle Korean lexical distinctions (e.g., 백부 vs.\ 숙부).

\paragraph{Cipher}
The English Hard tier operates on plaintexts of only $6\text{--}10$ characters, down from $20\text{--}24$ in Easy, yet reference accuracy drops by 52\,pp. At the same time, the Hard tier adds a positional transposition atop a Vigenère cipher. This pattern associates the observed tier ordering with algorithmic indirection rather than sequence length.

\paragraph{Saju and Time}
Instance length is invariant across tiers, yet reference accuracy falls from 79\% to 23\% for Saju and from 77\% to 19\% for Time. The tiers are instead constructed by reweighting the task mixture, replacing intermediate guidance with unassisted sexagenary ($60$-ganji) conversions.

These contrasts motivate calibrating difficulty against measured model performance rather than relying on input dimensions alone.

\subsection{Error Analysis: How Cipher Fails}
\label{sec:cipher-errors}

We analyze the $2{,}119$ incorrect, non-empty Korean Cipher outputs from 11 of the 12 non-floor models ($3{,}300$ total; $820$ correct and $361$ empty). We exclude Llama-4-Maverick because its $0.7\%$ English accuracy provides no evidence of Cipher competence. We decompose outputs and gold plaintexts into Hangul \emph{jamo} streams and compute similarity as one minus the Levenshtein distance \citep{levenshtein1966binary}, normalized by the length of the longer stream. Two baselines aid interpretation: pairs of unrelated gold plaintexts of equal length score $0.16$ because of skewed \emph{jamo} frequencies, while comparing ciphertext with plaintext scores $0.07$; direct echoes occur in at most $1\%$ of failures.

Models strong at Jamo Composition recover most plaintext \emph{jamo} even when final answers are wrong (GPT-5.5: $0.85$, $n{=}19$; Qwen3.5-397B: $0.89$). Models weak at it remain near the unrelated-text baseline ($0.11\text{--}0.21$ for Solar-100B, EXAONE-4.0, gpt-oss-120b, and Qwen3.5-27B). Similarity correlates with Jamo Composition accuracy ($r{=}0.75$, $p<.01$), with Qwen3.5-9B the sole exception ($0.90$ similarity despite $12.3\%$ accuracy). Dice block overlap yields the same pattern, and answer lengths match gold within one character. Re-segmenting a correct \emph{jamo} stream and applying a constant index shift both yield zero matches ($0\%$). Sub-syllabic competence therefore appears to shape whether Cipher succeeds and how it fails.

\paragraph{\jy{Operation-level probe.}}
\jy{To separate sub-syllabic operations from generic symbol manipulation, we factorize a 150-item subset (50 per tier) of the Jamo Composition items into four conditions: (i) shift-only on pre-decomposed \emph{jamo} tuples, (ii) decompose then shift, (iii) shift then recompose, and (iv) the full task. Across the 12 non-floor models, mean accuracy is $97.9$, $26.7$, $28.2$, and $21.9\%$, respectively. Shift-only accuracy is at least $95\%$ for 11 of 12 models (minimum $88\%$), including models at floor on Korean Cipher, whereas adding either decomposition or recomposition lowers accuracy for every model (exact McNemar, Holm-corrected; all 36 contrasts $p\le.004$). Korean Cipher accuracy correlates with conditions (ii)--(iv) ($r=.93$--$.95$; partial $r=.83$--$.89$ controlling for Direct-KO) but not with shift-only ($r=.39$, n.s.; partial $r=.02$). With $n=12$ we read these correlations cautiously, but they weaken a generic symbol-manipulation account and localize the shared difficulty to mapping between composed syllable blocks and their \emph{jamo}.}

\section{Conclusion}
\label{sec:conclusion}

We presented \textsc{Nolli}, a procedurally generated, difficulty-calibrated English--Korean puzzle benchmark with verified unique solutions and deterministic evaluation. Matched translations show small English--Korean differences, while script-adapted and Korean-only tasks reveal distinct model-specific patterns. Korean Cipher shows large drops for several models, unlike Cryptarithmetic over the same Hangul \emph{jamo}, and Jamo Composition is strongly associated with Korean Cipher performance. The Easy-to-Hard ordering also transfers broadly across models despite weak correspondence with structural size. \textsc{Nolli} supports controlled cross-lingual comparison alongside targeted evaluation of language-specific task demands.

\section*{Limitations}

\paragraph{Calibration and regeneration.}
Difficulty tiers are defined against one reference model at one reasoning-effort setting, so absolute band membership need not transfer to other models.
\jy{Recalibrating four representative conditions with two alternative references preserved tier ordering in 7 of 8 configurations, with Korean Cipher as the exception (Section~\ref{sec:calibration}); we therefore claim ordinal, not absolute, transfer across reference models, and only for this subset of conditions.}
Cipher and Cryptarithmetic are also calibrated independently by language \jy{and are not translation-equivalent, so their English--Korean gaps reflect different script-specific task demands rather than an effect of script alone}. A held-out check on six task types preserves tier ordering and most band memberships (Appendix~\ref{app:regeneration}), but full-suite stability remains unverified.

\paragraph{\jydel{The sub-syllabic mechanism is correlational.}\jy{The underlying mechanism remains unresolved.}}
\jydel{The Jamo Composition--Cipher relationship covers 12 models, but both tasks operate on \emph{jamo}. Without measuring per-model \emph{jamo} merging or including an English character-level control, we cannot separate a Korean-specific bottleneck from a general weakness in subword manipulation.}
\jy{The operation-level probe (Section~\ref{sec:cipher-errors}) localizes Korean Cipher difficulty to syllable decomposition and recomposition rather than to symbol shifting, but its correlations rest on 12 models and it does not identify why these operations fail. English Cipher serves only as a benchmark-level English-script counterpart; we include no item-matched English character-level control.}

\paragraph{Korean-only contrasts are task-confounded.}
Korean-only tasks differ from the Direct-KO baseline in task family, answer space, context length, and procedural demands\jy{, and Kinship additionally requires lexical knowledge of kinship terms that the prompt does not define}. Their deficits are therefore diagnostic rather than causal. Kinship is also the only cultural-lexicon task, limiting generalization to Korean cultural knowledge more broadly.

\paragraph{Scope.}
Our conclusions concern exact-match puzzle solving in one language pair under the configurations in Appendix~\ref{app:model-configs}, without partial credit. Equivalence is relative to a $\pm 10$\,pp margin and does not imply equal accuracy across languages.

\section*{\jy{Ethics Statement}}
\jy{All instances in \textsc{Nolli} are procedurally generated and contain no personal, sensitive, or copyrighted content.
Two of the cultural Korean-only tasks warrant comment: Saju uses the sexagenary calendar, which we treat as deterministic date arithmetic rather than as divination, and Kinship tests knowledge of Korean kinship terminology as lexical convention rather than as social norm.
Model evaluation used publicly available APIs and open-weight checkpoints under their respective licenses.
The generators, evaluators, and benchmark instances are released under the MIT License.}

\section*{Acknowledgments}

This work was supported by Korea Internet \& Security Agency (KISA) grant funded by the Korea government (PIPC) (No. RS-2026-25526342, Development of Technologies for Preventing Sensitive Information Inference and Risk Assessment in Foundation Model Operations). 
\\ This work was supported by Hankuk University of Foreign Studies Research Fund and by the National Research Foundation of Korea (NRF) grant funded by the Korea government (MSIT) (RS-2026-25606338). 
 \\ This work was supported by the Scaleup TIPS Program[RS-2026-25612218] funded by the Ministry of SMEs and Startups(MSS, Korea) in 2026
\\This research was supported by the ``Open Source AI·SW Developer and Community Support Program'' funded by the Ministry of Science and ICT (MSIT), Republic of Korea.
\\ This research was supported by the ``Advanced GPU Utilization Support Program'' funded by the Government of the Republic of Korea (Ministry of Science and ICT). 
\\This research was supported by the 'Open Source AI·SW Development and Utilization Support Program' through the Ministry of Science and ICT (MSIT) and the National IT Industry Promotion Agency (NIPA) (Grant Number PJT-26-040157).

\bibliography{custom}

\input{appendix}

\end{document}

%% file: appendix.tex
 \appendix
\newtcolorbox{taskexample}[1]{
  breakable, colback=gray!4, colframe=gray!45,
  title=#1, fonttitle=\bfseries\footnotesize,
  fontupper=\scriptsize, left=4pt, right=4pt, top=3pt, bottom=3pt,
  boxrule=0.4pt, arc=2pt, before skip=6pt, after skip=6pt
}




%

\section{Task Specifications}
\label{app:tasks}

This appendix specifies each of the 15 puzzle types: task statement, answer
format, difficulty levers, and a representative released instance. Per-tier
lever values appear in Table~\ref{tab:levers} and the accuracies they induce
in Appendix~\ref{app:calibration}.
In each box, \jydel{\emph{Puzzle} and \emph{Question} come from}\jy{\emph{Question} condenses} the instance's
\texttt{question} field \jy{(instructions summarized in English)}, with long tables, constraint lists, or dialogue turns
abbreviated as \texttt{[...]}; \jydel{\emph{Rules} paraphrases \mbox{Section~\ref{sec:tasks}} rather than the prompt, and} the evaluation-time system
prompt is omitted. Answers are the exact gold strings, so where an elision
removes content the answer depends on, the box illustrates the input and
answer formats rather than a hand-checkable derivation.

\begin{table*}[t]
\centering
\footnotesize
\setlength{\tabcolsep}{6pt}
\renewcommand{\arraystretch}{0.88}
\begin{tabular}{@{}llccc@{}}
\toprule
\textbf{Type} & \textbf{Lever} & \textbf{Easy} & \textbf{Medium} & \textbf{Hard} \\
\midrule
\multicolumn{5}{@{}l}{\emph{Direct translations (English and Korean share identical generator configurations)}} \\
Array Formula & product rows & 24--32 & 55--65 & 80--85 \\
 & order rows & 35--50 & 120--170 & 220--300 \\
 & template mix & med.\ .82 / easy .18 & hard .50 / med.\ .50 & hard only \\
 & distractor columns & 0 & +5 & +11 \\
Causal DAG & events & 25--31 & 40--46 & 46--58 \\
 & edge density & 0.62 & 0.88 & 0.95 \\
 & AND-dependency ratio & 0.47 & 0.63 & 0.64 \\
Inequality & grid size & $5{\times}5$ & $5{\times}5$ .62 / $6{\times}6$ .38 & $5{\times}5$ .20 / $6{\times}6$ .80 \\
Minesweeper & grid & $7{\times}7$--$12{\times}12$ & $9{\times}9$ & $9{\times}9$ \\
 & mines & 6--24 & 14 & 18 \\
 & revealed cells & fixed count & until unique ($\le$.45) & until unique ($\le$.58) \\
Number Baseball & digits & 7 & 8 & 8 \\
 & revealed digits & 5 & 0 & 0 \\
 & pinned positions & --- & 0.6 & --- \\
SAT Puzzle & variables & 9 & 11 & 14 \\
 & clauses & 36--54 & 56--88 & 115--170 \\
Sudoku & givens & 41 & 38 & 33 \\
 & queried cells & 4 & 5 & 6 \\
Yacht Dice & queried rounds & 2 & 2 & 8 \\
 & greedy--optimal gap & 0--8 & 7--30 & $\ge$30 \\
\midrule
\multicolumn{5}{@{}l}{\emph{Script adaptations (calibrated independently per language)}} \\
Cipher (EN) & cipher layers & 1 & 1 & 2 \\
 & answer length (chars) & 20--24 & 20--24 & 6--10 \\
 & keyword logic & direct & extraction & positional \\
Cipher (KO) & jamo cipher layers & 6 & 12 & 12 \\
 & answer length (chars) & 8 & 4--6$^{\dagger}$ & 6--8$^{\dagger}$ \\
 & keyword logic & positional & positional & extraction \\
Cryptarith.\ (EN) & operand length (letters) & 6 & 7 & 8 \\
 & minimum carries & 2 & 2 & 3 \\
Cryptarith.\ (KO) & operand length (jamo) & 6 & 7 & 7--8 \\
 & minimum final consonants & 2 & 2 & 3 \\
\midrule
\multicolumn{5}{@{}l}{\emph{Korean-only}} \\
Kinship & distractor utterances & 15 & 62 & 112 \\
 & relation hops & 4--5 & 3--4 & 2--3 \\
Saju & sub-type mix (med.\ : hard) & 4 : 1 & 2 : 3 & 1 : 8 \\
Time & \emph{ganji} fraction & 0.34 & 0.60 & 0.91 \\
Jamo Composition & syllables & 2 & 2 & 3 \\
 & final-consonant mode & light & single & mixed \\
Korean Units & summed items & 17 & 24 & 26 \\
 & maximum coefficient & 1{,}249 & 2{,}499 & 6{,}999 \\
\bottomrule
\end{tabular}
\caption{Per-tier difficulty-lever values, read from the released generator configurations. Ranges denote per-instance uniform sampling, except Korean Cipher answer length, which follows the plaintext word pool and skews short (Medium 75\% four-syllable; Hard 60\% six-syllable); ``---'' marks a parameter not set at that tier. Cipher and Cryptarithmetic are listed per language because their Korean variants are not translation-equivalent (Section~\ref{sec:crosslingual-design}).}
\label{tab:levers}
\end{table*}

\subsection{Direct Translations}
\label{app:tasks-direct}

For all eight types below, the Korean variant \jydel{is a faithful translation of the English variant generated from the identical algorithm and parameters}\jy{uses the identical algorithm and parameters as the English variant, rendered with author-written Korean templates}; only the prompt language differs.

\paragraph{Array Formula (\texttt{array\_formula}).}
Given several relational tables (e.g., products and orders) and a spreadsheet-style condition, the model must perform multi-step filtering, joining, and conditional aggregation to derive a single number.
\emph{Answer:} an integer.
\emph{Measures:} precise multi-step arithmetic and conditional aggregation over structured data.
\emph{Levers:} template mix; table size.
\begin{taskexample}{Array Formula}
\textbf{Question.} Given a Products table and an Orders table (columns abbreviated below; the released instances also include category, discount, supplier, warehouse, region, channel, priority columns, and a Customers table not needed for this query):
\begin{adjustbox}{max width=\linewidth}
\begin{BVerbatim}

[Products Table]
id | product   | price | stock | [...]
1  | Salmon    | 3200  | 109   | [...]
2  | Bread     | 1900  | 33    | [...]
[...]
65 | Makgeolli | 4400  | 10    | [...]
 
[Orders Table]
order_id | product  | quantity | quarter | [...]
ORD-001  | Lemon    | 44       | Q1      | [...]
ORD-002  | Mackerel | 1        | Q3      | [...]
[...]
ORD-131  | Orange   | 26       | Q4      | [...]
\end{BVerbatim}
\end{adjustbox}
\\
\\ What is the total revenue from orders where the quantity exceeds the average order quantity (25)? (revenue = price $\times$ quantity)
 
\textbf{Answer.} \texttt{7395000}
\end{taskexample}

\paragraph{Causal DAG (\texttt{causal\_dag}).}
Given a DAG of causal events with time-indexed trigger rules (AND/OR dependencies), the model propagates an initial condition through the graph and reports how many events fire.
\emph{Answer:} an integer (number of triggered events).
\emph{Measures:} conditional propagation and dependency tracking over graphs.
\emph{Levers:} number of events (25--31 $\to$ 46--58, sampled per instance); edge density (0.62$\to$0.95); AND-dependency ratio (0.47$\to$0.64).
\begin{taskexample}{Causal DAG}
\textbf{Question.} A system of causal events propagates over time: each event triggers its listed effects after a specified delay, an \textsc{OR} node fires as soon as its first prerequisite occurs, and an \textsc{AND} node fires only once all of its prerequisites have occurred.
\begin{adjustbox}{max width=\linewidth}
\begin{BVerbatim}
Events:
  E1: BackupFailed (Automated backup process fails)
  E10: CustomerComplaint (Support ticket is created)
  [...]
  E31: WindDamage (Strong winds cause infrastructure damage)
 
Causal Relationships (showing time delays):
  E27 (PowerSurge) -> E28 (QualityIssue): 38 minutes
  [E13 (ServerDown) OR E16 (CapacityReached) OR E17 (NetworkFailure)]
      -> E3 (Earthquake): 40 minutes         (Triggered by FIRST prerequisite)
  [E13 (ServerDown) AND E18 (PriceChanged) AND E3 (Earthquake)]
      -> E30 (InventoryLow): 36 minutes      (Requires ALL prerequisites)
  [...]
  E11 (TrafficJam) -> E17 (NetworkFailure): 12 minutes
 
Initial Condition:
- Event E1 (BackupFailed) occurs at minute 12
\end{BVerbatim}
\end{adjustbox}
By the time event E17 (NetworkFailure) first occurs, how many distinct listed events have occurred? (Include the initial event and the target event in the count.)
 
\textbf{Answer.} \texttt{19}
\end{taskexample}

\paragraph{Inequality (\texttt{inequality}).}
The model solves a Futoshiki puzzle: an $N{\times}N$ Latin square (each row and column holding $1$--$N$ exactly once), partially filled, subject to ``$<$''/``$>$'' constraints between orthogonally adjacent cells.
\emph{Answer:} the completed grid as $N^2$ space-separated numbers in row-major order.
\emph{Measures:} Latin-square constraint satisfaction under adjacency inequalities.
\emph{Levers:} grid size, shifting from $5{\times}5$ at Easy to an 80\% share of $6{\times}6$ at Hard.
\begin{taskexample}{Inequality}
\textbf{Question.} Fill the grid below so every row and column contains each number from 1 to 5 exactly once (a Latin square), subject to the ``$<$''/``$>$'' constraints between orthogonally adjacent cells, and provide the completed grid as 25 numbers, read row by row.
\begin{Verbatim}
Given grid (rows from top to bottom):
Row 1: 3 _ _ _ _
Row 2: _ _ _ _ _
Row 3: 5 _ _ _ _
Row 4: _ _ _ _ _
Row 5: _ _ _ _ _
 
Inequality constraints:
- (1,1) > (1,2)      - (1,1) > (2,1)      - (3,3) > (3,4)
- (3,4) > (4,4)       - (4,1) > (5,1)      - (4,2) < (4,3)
- (4,3) < (5,3)
\end{Verbatim}
 
\textbf{Answer.} \texttt{3 1 2 5 4 2 5 1 4 3 5 3 4 2 1 4 2 3 1 5 1 4 5 3 2}
\end{taskexample}

\paragraph{Minesweeper (\texttt{minesweeper}).}
Given a partially revealed Minesweeper grid (numbers and hidden cells), the model deduces the unique set of mine locations from local adjacency constraints.
\emph{Answer:} a list of coordinates.
\emph{Measures:} deduction over neighbor-count constraints (uniqueness guaranteed by the generator).
\emph{Levers:} grid size and mine density (Easy mixes blocks from $7{\times}7$/6 mines to $12{\times}12$/24 mines; Medium and Hard fix $9{\times}9$ at 14 and 18 mines).
\begin{taskexample}{Minesweeper}
\textbf{Question.} A $7{\times}7$ grid hides 6 mines; each revealed number (0--8) indicates how many of its 8 neighboring cells contain a mine, \texttt{\#} marks a hidden cell, and the puzzle has exactly one unique solution. Determine the exact location of all 6 mines, as (row, col) pairs sorted by row then column.
\begin{adjustbox}{max width=\linewidth}
\begin{BVerbatim}
   c0 c1 c2 c3 c4 c5 c6
r0 0  1  #  #  #  #  #
r1 0  #  2  #  2  #  2
r2 #  0  #  #  #  2  #
r3 #  #  1  2  #  #  1
r4 #  0  0  0  #  0  0
r5 0  0  0  #  #  1  #
r6 0  #  #  0  0  1  #
\end{BVerbatim}
\end{adjustbox}
 
\textbf{Answer.} \texttt{(0,2), (0,6), (2,3), (2,4), (2,6), (6,6)}
\end{taskexample}

\paragraph{Number Baseball (\texttt{number\_baseball}).}
From Strike/Ball feedback on a series of guesses (Bulls \& Cows), the model infers the hidden $N$-digit number with all-distinct digits.
\emph{Answer:} an $N$-digit number.
\emph{Measures:} integrating multiple feedback constraints into a single deduction.
\emph{Levers:} number of digits (7$\to$8); revealed helper digits (5$\to$0) and pinned positions, which Medium alone retains.
\begin{taskexample}{Number Baseball}
\textbf{Question.} The secret number has 7 digits, all distinct. For each guess below, ``Strike'' counts digits correct in both value and position, and ``Ball'' counts digits correct in value but in the wrong position. Find the unique 7-digit secret number satisfying all hints.
\begin{Verbatim}
Hints:
  1. Guess: 3280167 -> 2 Strike(s), 2 Ball(s)
  2. Guess: 0461329 -> 2 Strike(s), 3 Ball(s)
  3. Guess: 7253698 -> 1 Strike(s), 5 Ball(s)
  4. Guess: 0943176 -> 1 Strike(s), 4 Ball(s)
  5. Guess: 7380529 -> 2 Strike(s), 3 Ball(s)
 
Known digits (these 5 digits appear in the secret, positions unknown;
the other 2 are for you to deduce): 2, 4, 5, 7, 9
\end{Verbatim}
 
\textbf{Answer.} \texttt{3264579}
\end{taskexample}

\paragraph{SAT Puzzle.}
Boolean constraints stated in natural language (e.g., ``at least one of \dots is telling the truth'') over a guilty/innocent scenario; the model must produce a satisfying truth assignment.
\emph{Answer:} a variable$\to$\texttt{True}/\texttt{False} assignment.
\emph{Measures:} propositional satisfiability.
\emph{Levers:} number of variables (9$\to$14); number of clauses.
\begin{taskexample}{SAT Puzzle}
\textbf{Question.} A crime has been committed; each of the 9 suspects below is either guilty or innocent, and every constraint (each an ``at least one of ... is true'' disjunction) must hold simultaneously. Who is guilty and who is innocent?
\begin{adjustbox}{max width=\linewidth}
\begin{BVerbatim}
Variables: Alice, Leo, Emma, David, Nick, Carol, Bob, Iris, Mary
 
Constraints:
  1. At least one of the following is true: Carol is guilty, or David is
     innocent, or Emma is innocent, or Alice is guilty
  2. At least one of the following is true: Alice is innocent, or Nick is
     guilty, or David is guilty, or Carol is innocent
  [...]
  49. At least one of the following is true: Nick is innocent, or Mary is
     innocent, or Leo is innocent
\end{BVerbatim}
\end{adjustbox}
 
\textbf{Answer.} \texttt{\{"Alice": false, "Bob": true, "Carol": true, "David": true, "Emma": true, "Iris": true, "Leo": true, "Mary": false, "Nick": true\}}
\end{taskexample}

\paragraph{Sudoku.}
The model completes a partially filled 9$\times$9 Sudoku and reports the values of specific queried cells.
\emph{Answer:} the queried cell values in order.
\emph{Measures:} constraint-satisfaction completion plus targeted extraction.
\emph{Levers:} number of givens (41$\to$33).

\begin{taskexample}{Sudoku}
\textbf{Question.} Fill the $9{\times}9$ grid below so that each row, column, and $3{\times}3$ box contains digits 1--9 exactly once, then report the values at (row, col) coordinates (5,4), (4,2), (6,2), (1,7), in that order.
\begin{Verbatim}
. . 2 6 8 . 4 . 5
1 . . . . . . 7 .
7 6 . . 4 . 2 3 9
. 3 . 7 5 4 9 . .
. 7 . 1 . 2 . 8 .
. . 9 3 6 8 . 4 .
9 4 1 . 3 . . 5 8
. 8 . . . . . . 2
6 . 3 . 7 5 1 . .
\end{Verbatim}
 
\textbf{Answer.} \texttt{1 3 1 4}
\end{taskexample}

\paragraph{Yacht Dice.}
Given twelve rounds of dice results, the model assigns each round to a scoring category (full house, straight, yacht, etc.)\ under the game's rules and computes the resulting score.
\emph{Answer:} an integer score.
\emph{Measures:} combinatorial assignment plus exact rule application.
\emph{Levers:} number of queried rounds (2$\to$8) and the greedy--optimal score gap band; the 12-round horizon and dice count are fixed.

\begin{taskexample}{Yacht Dice}
\textbf{Question.} Twelve rounds of five-dice rolls below must each be assigned to a distinct scoring category (Full House 25, Small Straight 30, Large Straight 40, Yacht 50, Aces--Sixes = sum of matching dice, upper-section bonus +35 if the Aces--Sixes total is 63 or more, etc.). Compute the optimal 12-round assignment that maximizes the total score, then report only the sum of the per-round scores for rounds 4 and 7 (not the full 12-round total).
\begin{Verbatim}
Round 1: [4, 1, 2, 1, 2]      Round 2: [3, 5, 1, 3, 5]
Round 3: [4, 1, 3, 3, 1]      Round 4: [5, 1, 1, 1, 5]
[...]
Round 11: [5, 6, 6, 2, 2]     Round 12: [6, 3, 4, 4, 6]
\end{Verbatim}
 
\textbf{Answer.} \texttt{42}
\end{taskexample}

\subsection{Script Adaptations}
\label{app:tasks-script}

Both types exist in English and Korean, but the Korean variants are adapted to the compositional jamo structure of Hangul and are therefore not translation-equivalent to their English counterparts.

\paragraph{Cipher.}
The model decrypts a ciphertext by following an explicitly specified stack of classical ciphers (e.g., Vigen\`ere) with given keywords.
The English variant operates over the Roman alphabet; the Korean variant operates over jamo indices (initial/medial/final components of each syllable block).
\jy{Korean Easy plaintexts follow a fixed eight-syllable, four-slot template (about $2.7{\times}10^{4}$ possible strings), whereas English Easy plaintexts are random 20--24-letter strings. Korean Medium and Hard plaintexts are drawn from fixed pools of 306 and 291 compound nouns (4--6 and 6--8 syllables).}
\emph{Answer:} the plaintext string.
\emph{Measures:} faithful character-level transformation from a specification.
\emph{Levers:} cipher stack depth; key-derivation complexity.

\begin{taskexample}{Cipher (English)}
\textbf{Question.} A Vigen\`ere cipher adds a repeating keyword to the plaintext letters (A=0, B=1, ...); one worked example using the same keyword and algorithm is given as a hint. Decrypt the ciphertext (uppercase, no spaces).
\begin{Verbatim}
Ciphertext: 'SCTAHLOTYJOUZEPPHVUAMAE'
Keyword: 'FOXTROT'
Applied algorithm: VIGENERE
Worked example: IRRVL -> NFOOC
\end{Verbatim}
 
\textbf{Answer.} \texttt{NOWHQXVOKMVDLLKBKCDMTVQ}
\end{taskexample}

\begin{taskexample}{Cipher (Korean)}
\textbf{Question.} Each Hangul syllable is decomposed into initial/medial/final \emph{jamo}. \texttt{CHO\_SHIFT} adds the keyword's initial-consonant indices (mod 19) to the plaintext's initial consonants; \texttt{JUNG\_SUB} substitutes medial-vowel indices through a keyword-derived table; \texttt{REVERSE} reverses the whole string. The keyword is given only by its word position in the cover story below. Return the plaintext in Hangul without spaces.
\begin{Verbatim}
Cover story: "...키워드 달빛 부수어짐 없이 적용됨..."
  (keyword = the 6th word of this log excerpt)
Ciphertext: '텨쓰흐폔망브졔뙨'
Applied algorithm: CHO_SHIFT -> JUNG_SUB -> REVERSE ->
                   CHO_SHIFT -> JUNG_SUB -> CHO_SHIFT
\end{Verbatim}
 
\textbf{Answer.} \texttt{문화시장관리지도}
\end{taskexample}

\paragraph{Cryptarithmetic.}
Letters are mapped to digits through \emph{independent per-group tables} (alphabet groups in English; jamo groups in Korean), and the mapping must satisfy a given arithmetic equation.
\emph{Answer:} a numeric string.
\emph{Measures:} injective-mapping constraints combined with arithmetic satisfaction.
\emph{Levers:} operand length (EN: 6$\to$8 letters; KO: 6$\to$7--8 jamo); minimum carries (2$\to$3). The operand count is fixed at four in both languages.

\begin{taskexample}{Cryptarithmetic (English)}
\textbf{Question.} Each letter below maps to a digit (0--9). Letters are split into three independent groups by alphabet range (A--I, J--R, S--Z); within a group, different letters take different digits, but the same \jydel{letter may repeat across groups with unrelated values}\jy{digit may be reused across groups}, and no word may start with 0. Find the numeric value of OCTOPUS.
\begin{Verbatim}
Revealed: [A-I] A=0, B=5, E=1   [J-R] K=1, R=4   [S-Z] S=4
 
  BANANA
+ MARKET
+ GARDEN
+ MIRROR
---------
= OCTOPUS
\end{Verbatim}
 
\textbf{Answer.} \texttt{3203254}
\end{taskexample}

\begin{taskexample}{Cryptarithmetic (Korean)}
\textbf{Question.} Each Hangul syllable below decomposes into initial/medial/final \emph{jamo}; a syllable's numeric value concatenates its jamo's digits in that order (2 digits if there is no final consonant, 3 if there is). Initial consonants, medial vowels, and final consonants each form independent digit tables, and no word's first initial consonant may be 0. Find the numeric value of 얼굴.
\begin{Verbatim}
Revealed: [cho] g=0, m=3, b=2, o=9, j=4
          [jung] eo=3, yeo=7, wo=1
          [jong] g=3, o=9
 
  무지개
+ 병원
+ 선물
+ 식당
-----
= 얼굴
\end{Verbatim}
 
\textbf{Answer.} \texttt{930020}
\end{taskexample}

\subsection{Korean-Only Tasks}
\label{app:tasks-koonly}

\paragraph{Kinship.}
A multi-party conversation around a family photo mixes genuine cues (siblings, children, degrees of kinship) with distractor small talk; the model must resolve references across speakers and identify a target person or their correct Korean kinship term.
\emph{Answer:} a candidate label corresponding to a kinship term.
\emph{Measures:} knowledge of the Korean kinship-term system plus multi-party reference resolution.
\emph{Levers:} number of distractor utterances (15$\to$62$\to$112); relation-chain length, which \emph{decreases} from 4--5 hops at Easy to 2--3 at Hard (Section~\ref{sec:difficulty-mechanisms}).

\begin{taskexample}{Kinship}
\textbf{Question.} A multi-party conversation below around a family photo mixes genuine kinship cues with distractor small talk. Tracing the chain of relations across speakers, 나는 빨간 옷 입은 분은 어떻게 불러야 하는가? (Choose from a list of 26 candidate kinship terms.)
\begin{adjustbox}{max width=\linewidth}
\begin{BVerbatim}
준용: "뒤에 계신 분은 검은 정장 입은 분이랑 어떤 관계야?"
나: "검은 정장 입은 분은 뒤에 계신 분의 아들이야."
수용: "뒤에 계신 분한테 형제 있어? 사진 있어?"
나: "응, 여기 맨 앞에 계신 분이 여동생이야."
[... additional distractor and relation-tracing turns ...]
철수: "뒤에 계신 분한테 형제 있어?"
나: "응, 뒤에 계신 분은 긴 머리 분의 형이야."
다솔: "긴 머리 분은 결혼했어?"
나: "응, 빨간 옷 입은 분의 남편이 긴 머리 분이야."
\end{BVerbatim}
\end{adjustbox}
 
\textbf{Answer.} \texttt{F} (시숙모)
\end{taskexample}

\paragraph{Saju.}
Given a day pillar (stem--branch pair) and a birth hour, the model applies the traditional hour-pillar derivation rule to compute elements of the four-pillars calendar.
\emph{Answer:} a two-character stem--branch pair.
\emph{Measures:} rule application within the Korean sexagenary calendar system.
\emph{Levers:} problem-type mix.

\begin{taskexample}{Saju}
\textbf{Question.} 사주에서 일주(日柱)가 '임자'인 사람이 13시에 태어났다. 전통적인 시두법(五鼠遁) 규칙(일간에 따라 시주를 도출하는 규칙)을 적용한 시주(時柱) 간지는?
 
\textbf{Answer.} \texttt{정미}
\end{taskexample}

\paragraph{Time.}
Relative date statements (``today is the birthday,'' ``$X$ days later'') must be resolved to a Gregorian date, in some instances requiring the sexagenary day-cycle (일진, \emph{iljin}) to be computed from scratch.
\emph{Answer:} a date in \texttt{YYYY.M.D} format.
\emph{Measures:} date arithmetic plus Korean calendar knowledge.
\emph{Levers:} fraction of instances requiring sexagenary-cycle computation (0.34$\to$0.91).

\begin{taskexample}{Time}
\textbf{Question.} 1999년 새해 첫날에 "금일이 내 생일이야"라는 말을 들었다. 그 생일로부터 23일 후, 그 날의 양력 날짜는 무엇인가? ('YYYY.M.D' 형식으로 답하라.)
 
\textbf{Answer.} \texttt{1999.1.24}
\end{taskexample}

\paragraph{Jamo Composition.}
Each Hangul syllable is decomposed into its initial, medial, and final jamo; the initial consonants are cyclically shifted by a specified offset; and the syllables are recomposed.
\emph{Answer:} the transformed string.
\emph{Measures:} decomposition and manipulation of the Korean writing system---a purely orthographic task that cannot be translated.
\emph{Levers:} number of syllables; final-consonant complexity.

\begin{taskexample}{Jamo Composition}
\textbf{Question.} Each Hangul syllable decomposes into initial/medial/final \emph{jamo}; the initial consonant is cyclically shifted by 1 position within a 19-symbol order (wrapping from the last back to the first), while the medial and final components are left unchanged, and the syllable is then recomposed. Given the letters '닇데', what is the transformed string?
 
\textbf{Answer.} \texttt{딓떼}
\end{taskexample}

\paragraph{Korean Units (\texttt{korean\_units}).}
Several quantities in traditional Korean units must be converted to a base unit \emph{using only a conversion table supplied in the prompt}, scaled and signed as instructed, and summed.
Because the (randomized) table is self-contained, the task measures table-based multi-step conversion arithmetic rather than recall of real-world unit values.
\emph{Answer:} a single integer in the base unit (e.g., \texttt{465373918}평).
\emph{Measures:} multi-step conversion arithmetic over an in-context table.
\emph{Levers:} number of summed items; coefficient magnitude; conversion ratios.

\begin{taskexample}{Korean Units}
\textbf{Question.} Using only the self-contained conversion table below (traditional Korean area units), convert each listed item to the base unit '평', scale it by its given multiplier, and add or subtract it from a running total as indicated, then report the final sum.
\begin{adjustbox}{max width=\linewidth}
\begin{BVerbatim}
[넓이 환산표]
1단 = 19평   1마지기 = 7단   1정보 = 12마지기   1결 = 16정보
 
갑 [더함, 6배]: 287결 1034정보 286마지기 578단 195평
을 [더함, 7배]: 1147결 207정보 677마지기 967평
[...]
제17호 [더함, 6배]: 987결 277마지기 722단 492평
\end{BVerbatim}
\end{adjustbox}
 
\textbf{Answer.} \texttt{465373918평}
\end{taskexample}

\section{Construction of Direct-Translation Tasks}
\label{app:direct-construction}

The eight direct-translation types use language-independent procedural generators and fixed, author-written English and Korean rendering templates. No machine-translation service or LLM is invoked during generation. Depending on the task, the two language conditions either render the same latent instance (the three item-paired types) or sample independently from the same generator and parameter configuration; all eight are therefore matched in distribution. Because each language-specific template is reused across instances, linguistic validation was conducted at the template level rather than separately for all generated items. We checked the template pairs for semantic correspondence, placeholder coverage, answer-format consistency, and the absence of language-specific changes to the underlying constraints or scoring logic.

The cross-model TOST provides a complementary behavioral check: it finds no systematic accuracy difference exceeding the prespecified $\pm 10$\,pp margin. We treat this result as evidence against a substantial difficulty shift between the two rendered conditions, not as a substitute for linguistic validation.

\section{Calibration Details}
\label{app:calibration}

Table~\ref{tab:calibration} reports reference-model accuracy for all 75 task--tier combinations produced by the calibration protocol of Section~\ref{sec:calibration}.
Of the 51 directly calibrated tiers (24 English direct-translation, 12 script-adaptation, 15 Korean-only), every tier falls inside its target band and adjacent tiers are separated by at least 10\,pp.
The 24 mirrored Korean tiers of the direct-translation tasks inherit their English-calibrated generator configuration without re-tuning, and three of them exceed their band by 2--7\,pp.
Per-tier generator parameter values for every task are listed in Table~\ref{tab:levers}.

\begin{table}[t]
\centering
\footnotesize
\setlength{\tabcolsep}{3.2pt}
\renewcommand{\arraystretch}{0.92}
\begin{tabular}{@{}l @{\hspace{7pt}} ccc @{\hspace{9pt}} ccc@{}}
\toprule
& \multicolumn{3}{c}{\textbf{English}} & \multicolumn{3}{c}{\textbf{Korean}} \\
\cmidrule(lr){2-4} \cmidrule(l){5-7}
\textbf{Task} & Easy & Med. & Hard & Easy & Med. & Hard \\
\midrule
\multicolumn{7}{@{}l}{\emph{Direct translations (KO mirrored)}} \\
Array Formula & 74 & 50 & 31 & 69 & 52 & 23 \\
Causal DAG & 79 & 57 & 32 & 78 & 53 & 35 \\
Inequality & 77 & 58 & 21 & 73 & 50 & 20 \\
Minesweeper & 76 & 44 & 22 & 77 & 45 & 17 \\
Number Baseball & 82 & 53 & 32 & 85 & 57 & 42$^{*}$ \\
SAT Puzzle & 79 & 53 & 17 & 87$^{*}$ & 52 & 22 \\
Sudoku & 81 & 52 & 23 & 82 & 58 & 17 \\
Yacht Dice & 76 & 58 & 31 & 77 & 51 & 38$^{*}$ \\
\addlinespace[2.5pt]
\multicolumn{7}{@{}l}{\emph{Script adaptations (calibrated per language)}} \\
Cipher & 75 & 55 & 23 & 79 & 50 & 35 \\
Cryptarithmetic & 78 & 47 & 21 & 85 & 54 & 26 \\
\addlinespace[2.5pt]
\multicolumn{7}{@{}l}{\emph{Korean-only}} \\
Kinship & --- & --- & --- & 77 & 46 & 31 \\
Saju & --- & --- & --- & 79 & 50 & 23 \\
Time & --- & --- & --- & 77 & 54 & 19 \\
Jamo Composition & --- & --- & --- & 69 & 50 & 33 \\
Korean Units$^{\dagger}$ & --- & --- & --- & 76 & 58 & 34 \\
\bottomrule
\end{tabular}
\caption{Reference-model accuracy (\%) for all 75 task--language--tier cells
(\texttt{gemini-3-flash}, reasoning effort \emph{medium}, $n{=}100$ per tier;
target bands 75/50/25\,$\pm$10\,pp). ---: no English counterpart.
$^{*}$Above band (3 of 75, all mirrored Korean tiers, by 2--7\,pp). $^{\dagger}$Separate calibration run, identical protocol.}
\label{tab:calibration}
\end{table}

\jy{Table~\ref{tab:ref-sensitivity} details the reference-model sensitivity analysis of Section~\ref{sec:calibration}.}

\begin{table}[t]
\centering
\footnotesize
\setlength{\tabcolsep}{4pt}
\begin{tabular}{@{}lccc@{}}
\toprule
& \multicolumn{3}{c}{\textbf{Evaluated by}} \\
\cmidrule(l){2-4}
\textbf{Condition} & \textbf{Gemini} & \textbf{Qwen} & \textbf{DeepSeek} \\
\midrule
\multicolumn{4}{@{}l}{\emph{Calibrated on Qwen3.5-27B}} \\
Causal DAG EN & 79/57/32 & \textbf{74/46/27} & 64/38/19 \\
Cipher EN & 82/80/36 & \textbf{73/55/22} & 93/91/49 \\
Cipher KO$^{*}$ & 97/83/48 & \textbf{75/53/25} & 82/55/28 \\
Time KO & 77/68/44 & \textbf{67/51/27} & 69/55/28 \\
\addlinespace[2.5pt]
\multicolumn{4}{@{}l}{\emph{Calibrated on DeepSeek-V4-Flash}} \\
Causal DAG EN & 90/75/32 & 88/68/27 & \textbf{70/57/23} \\
Cipher EN & 73/11/6 & 47/15/14 & \textbf{82/55/21} \\
Cipher KO & 98/95/35 & \textit{11/18/0} & \textbf{73/60/20} \\
Time KO & 77/54/43 & 67/37/21 & \textbf{65/42/29} \\
\bottomrule
\end{tabular}
\caption{\jy{Cross-reference evaluation of recalibrated configurations (Easy/Medium/Hard accuracy, \%, $n{=}100$ per tier). Gemini = Gemini 3 Flash, Qwen = Qwen3.5-27B, DeepSeek = DeepSeek-V4-Flash. Bold: the calibrating reference, which falls in the target bands (65--85/40--60/15--35). Only tiers outside the bands for that reference were recalibrated. Easy$>$Medium$>$Hard holds for every evaluator except the italic cell, where Qwen3.5-27B is near floor on DeepSeek-calibrated Korean Cipher. $^{*}$Qwen3.5-27B solved none of the original Korean Cipher items (0/0/0\%), so its tiers mix original items with reverse-only items (80/54/27\% of Easy/Medium/Hard), which keep the tier's configuration but replace the cipher stack with a single REVERSE step; Qwen3.5-27B solved 92--98\% of reverse-only items. Decoding: Gemini 3 Flash temperature 1.0, top-$p$ 0.95, top-$k$ 64, reasoning effort medium; Qwen3.5-27B temperature 0.6, top-$p$ 0.95, top-$k$ 20, default reasoning; DeepSeek-V4-Flash temperature 1.0, top-$p$ 1.0, default reasoning. These configurations serve this analysis only; the released benchmark uses the Gemini-calibrated configurations.}}
\label{tab:ref-sensitivity}
\end{table}

\section{Held-Out Regeneration Check}
\label{app:regeneration}

To assess whether calibrated difficulty persists beyond the released instances, we conduct a held-out regeneration check using fresh random seeds. For each tier, we generate 50 instances from the same calibrated generator configuration and evaluate them with the reference model, Gemini~3 Flash with reasoning effort set to \emph{medium}. 
We evaluate six representative puzzle types: Cipher and Cryptarithmetic in both English and Korean, Array Formula and Causal DAG in English, and Jamo Composition and Time in Korean. This yields eight language-specific tasks and 24 task--tier cells.

All 16 adjacent-tier comparisons satisfy the required separation criterion, with Easy--Medium and Medium--Hard gaps of at least 10 percentage points. Twenty of the 24 evaluated cells (83\%) remain within their target bands. The four exceptions miss their nearest band boundary by 1--7\,pp and do not alter the Easy $>$ Medium $>$ Hard ordering. Given the $n=50$ sample size per cell, these modest shifts are plausibly attributable to sampling variation. Within this representative subsample, regenerated instances therefore retain the calibrated ordering and approximate absolute difficulty, although full-suite stability remains untested.

\begin{table}[t]
\centering
\footnotesize
\setlength{\tabcolsep}{1.5pt}
\renewcommand{\arraystretch}{0.92}
\begin{tabular}{lcccccc}
\toprule
& \multicolumn{3}{c}{English} & \multicolumn{3}{c}{Korean} \\
\cmidrule(lr){2-4}\cmidrule(l){5-7}
Task & Easy & Medium & Hard & Easy & Medium & Hard \\
\midrule
\multicolumn{7}{l}{\textit{Script adaptations}} \\
Cipher
& 70 & 60 & 28
& 82 & 56 & 30 \\
Cryptarithmetic
& 84 & 48 & 14$^{*}$
& 92$^{*}$ & 62$^{*}$ & 26 \\
\midrule
\multicolumn{7}{l}{\textit{Direct translations}} \\
Array Formula
& 84 & 52 & 40$^{*}$
& --- & --- & --- \\
Causal DAG
& 76 & 46 & 30
& --- & --- & --- \\
\midrule
\multicolumn{7}{l}{\textit{Korean-only}} \\
Jamo Composition
& --- & --- & ---
& 82 & 44 & 20 \\
Time
& --- & --- & ---
& 76 & 48 & 30 \\
\bottomrule
\end{tabular}
\caption{Held-out regeneration accuracy (\%) on 50 fresh instances per tier,
generated from unseen seeds with the calibrated configurations and reference
model of Table~\ref{tab:calibration}. ``---'': language condition not in this
subsample. $^{*}$Outside the target band, by at most 7\,pp.}
\label{tab:regeneration}
\end{table}

\section{Equivalence Testing Details}
\label{app:equivalence}

We apply two one-sided tests (TOST) for two independent samples with a $\pm 10$\,pp equivalence margin, matching the tolerance of the difficulty bands (Section~\ref{sec:calibration}).
Equivalence holds for all 12 non-floor models (max Holm-corrected $p = 3.1 \times 10^{-4}$).
As a robustness check, we restrict the analysis to the three tasks whose Korean and English items share identical random seeds (Causal DAG, Inequality, Minesweeper; $n{=}900$ paired items per model) and apply a paired TOST, which reproduces equivalence for all 12 models (max Holm-corrected $p = 2.5 \times 10^{-3}$).
Paired McNemar tests detect a point difference for four models (DeepSeek-V4-Flash, Gemini 3.1, Qwen3.5-27B, and EXAONE-4.0; all $p < .05$), but in each case the effect ($\le 4.8$\,pp) falls well within the equivalence margin: statistically distinguishable, practically equivalent.
Per-model gaps and test statistics appear in Table~\ref{tab:tost}.


\begin{table*}[t]
\centering
\small
\renewcommand{\arraystretch}{0.92}
\setlength{\tabcolsep}{5pt}
\begin{tabular}{@{}l ccc c @{\hspace{10pt}} ccc@{}}
\toprule
& \multicolumn{4}{c}{\textbf{Two-sample} (Direct, 8 tasks, $n{=}2{,}400$/lang)} & \multicolumn{3}{c}{\textbf{Paired} (3 seed-matched tasks, $n{=}900$)} \\
\cmidrule(lr){2-5} \cmidrule(l){6-8}
\textbf{Model} & EN & KO & $\Delta$ & TOST $p$ & $\Delta$ & McNemar $p$ & TOST $p$ \\
\midrule
Opus 4.8 & 79.6 & 77.4 & $+2.2$ & $2{\times}10^{-10}$ & $+0.8$ & 0.51 & $5{\times}10^{-19}$ \\
GPT-5.5 & 83.4 & 82.3 & $+1.0$ & $1{\times}10^{-15}$ & $+1.7$ & 0.08 & $7{\times}10^{-20}$ \\
Gemini 3.1 & 75.8 & 73.7 & $+2.1$ & $6{\times}10^{-10}$ & $+2.4$ & \textbf{0.04} & $5{\times}10^{-11}$ \\
Qwen3.5-9B & 39.8 & 38.8 & $+1.0$ & $2{\times}10^{-10}$ & $+2.1$ & 0.29 & $3{\times}10^{-5}$ \\
Qwen3.5-27B & 58.3 & 53.5 & $+4.8$ & $3{\times}10^{-4}$ & $+3.3$ & \textbf{0.04} & $3{\times}10^{-5}$ \\
Gemma-4-31B & 56.3 & 55.4 & $+0.9$ & $6{\times}10^{-10}$ & $+2.0$ & 0.19 & $1{\times}10^{-7}$ \\
gpt-oss-120b & 56.6 & 55.1 & $+1.5$ & $4{\times}10^{-9}$ & $-0.6$ & 0.79 & $1{\times}10^{-7}$ \\
DS-V4-Flash & 49.5 & 44.5 & $+5.0$ & $3{\times}10^{-4}$ & $+4.8$ & \textbf{0.01} & $2{\times}10^{-3}$ \\
Qwen3.5-397B & 68.3 & 69.5 & $-1.2$ & $2{\times}10^{-10}$ & $-0.7$ & 0.76 & $5{\times}10^{-7}$ \\
L4-Maverick & 20.6 & 18.9 & $+1.7$ & $4{\times}10^{-12}$ & $+0.2$ & 0.94 & $9{\times}10^{-11}$ \\
EXAONE-4.0 & 30.9 & 30.2 & $+0.7$ & $1{\times}10^{-11}$ & $-3.2$ & \textbf{0.04} & $1{\times}10^{-5}$ \\
Solar-100B & 14.7 & 17.4 & $-2.7$ & $3{\times}10^{-11}$ & $-2.3$ & 0.09 & $1{\times}10^{-8}$ \\
\bottomrule
\end{tabular}
\caption{Per-model equivalence statistics for the presentation-language (Direct) tasks. $\Delta = \text{EN}-\text{KO}$ (pp). TOST $p$-values are Holm-corrected across the 12 non-floor models; equivalence at the $\pm10$\,pp margin holds for every model under both the two-sample test over all 8 Direct tasks (max $p{=}3.1{\times}10^{-4}$) and the paired test over the three tasks whose English and Korean items share random seeds (max $p{=}2.5{\times}10^{-3}$). Uncorrected McNemar tests detect accuracy differences for four models ($p$-values in bold if $<.05$), but every such effect is $\le4.8$\,pp---within the equivalence margin.}
\label{tab:tost}
\end{table*}

\section{Model Configurations}
\label{app:model-configs}

We evaluate 15 models in three groups.
\emph{Frontier API models}: Claude Opus 4.8 \citep{anthropic2026claudeopus48}, GPT-5.5 \citep{openai2026gpt55}, and Gemini 3.1 Pro preview \citep{google2026gemini31}.
\emph{International open-weight models}: Llama-3.1-8B-Instruct \citep{grattafiori2024llama}, Qwen3.5-9B, Qwen3.5-27B, and Qwen3.5-397B-A17B (17B active) \citep{qwenteam2026qwen35}, Gemma-4-31B-it \citep{gemmateam2026gemma4technicalreport}, gpt-oss-120b (117B, 5B active) \citep{agarwal2025gpt}, DeepSeek-V4-Flash (284B, 13B active) \citep{xu2026deepseek}, and Llama-4-Maverick (400B, 17B active) \citep{meta2025llama4}.
\emph{Korean-developed models}: EXAONE-3.5-7.8B \citep{an2024exaone}, EXAONE-4.0-32B \citep{bae2025exaone}, Mi:dm-2.0-Base-Instruct (11.5B) \citep{shin2026mi}, and Solar-Open-100B (102B, 12B active) \citep{park2026solaropentechnicalreport}.

Table~\ref{tab:model-configs} summarizes the inference configurations used for
the reported results. Models are grouped by category; Gemini~3 Flash
is listed separately as the calibration reference only.

\begin{table*}[t]
\centering
\small
\renewcommand{\arraystretch}{0.93}
\setlength{\tabcolsep}{5pt}
\begin{tabular}{lllrl}
\toprule
Model & Inference Backend & Temperature & Max Output Tokens & Reasoning \\
\midrule
\multicolumn{5}{l}{\textit{Frontier API models}} \\
GPT-5.5 & OpenAI API & not sent & 32{,}768 & \texttt{reasoning\_effort=medium} \\
Claude Opus 4.8 & Anthropic API & not sent & 32{,}768 & \texttt{reasoning\_effort=medium} \\
Gemini 3.1 Pro & Google API & 1.0 & 32{,}768 & \texttt{reasoning\_effort=medium} \\
\midrule
\multicolumn{5}{l}{\textit{International open-weight models}} \\
Llama-3.1-8B & vLLM & 0.6 & 32{,}768 & reasoning off \\
Qwen3.5-9B & vLLM & 0.6 & 32{,}768 & reasoning off (non-thinking) \\
Qwen3.5-27B & OpenRouter & 0.6 & 32{,}768 & n/a \\
Qwen3.5-397B & vLLM & 0.6 & 32{,}768 & reasoning off (non-thinking) \\
Gemma-4-31B & vLLM & 1.0 & 32{,}768 & reasoning on \\
gpt-oss-120b & vLLM & 1.0 & 32{,}768 & \texttt{reasoning\_effort=medium} \\
DeepSeek-V4-Flash & OpenRouter & 1.0 & 32{,}768 & n/a \\
Llama-4-Maverick & vLLM & 0.6 & 32{,}768 & reasoning off \\
\midrule
\multicolumn{5}{l}{\textit{Korean-developed models}} \\
EXAONE-3.5 & vLLM & 0.0 & 14{,}336 & reasoning off \\
EXAONE-4.0 & vLLM & 0.6 & 32{,}768$^\dagger$ & reasoning on \\
Mi:dm-2.0 & vLLM & 0.8 & 14{,}336 & n/a \\
Solar-100B & vLLM & 0.8 & 32{,}768 & n/a \\
\midrule
\multicolumn{5}{l}{\textit{Calibration reference}} \\
Gemini 3 Flash & Google API & 1.0 & 32{,}768 & \texttt{reasoning\_effort=medium} \\
\bottomrule
\end{tabular}
\caption{Inference configurations for the reported results. ``not sent'': the reasoning API rejects an explicit temperature. ``n/a'': the serving interface exposes no reasoning control. $^\dagger$EXAONE-4.0 used \texttt{max\_tokens}=64{,}000 for 18 of 75 task--tier evaluations.}
\label{tab:model-configs}
\end{table*}

%
%

\section{Full Per-Task Results}
\label{app:full-results}

Table~\ref{tab:full-results} reports accuracy (\%) for every task, language, difficulty tier, and evaluated model.

\begin{sidewaystable*}[p]
\centering
\scriptsize
\setlength{\tabcolsep}{1.5pt}
\begin{tabular}{@{}ll*{45}{c}@{}}
\toprule
 & & \multicolumn{3}{c}{\rotatebox{90}{GPT-5.5}} & \multicolumn{3}{c}{\rotatebox{90}{Opus 4.8}} & \multicolumn{3}{c}{\rotatebox{90}{Gemini 3.1}} & \multicolumn{3}{c}{\rotatebox{90}{Qwen3.5-9B}} & \multicolumn{3}{c}{\rotatebox{90}{Qwen3.5-27B}} & \multicolumn{3}{c}{\rotatebox{90}{Qwen3.5-397B}} & \multicolumn{3}{c}{\rotatebox{90}{Gemma-4-31B}} & \multicolumn{3}{c}{\rotatebox{90}{gpt-oss-120b}} & \multicolumn{3}{c}{\rotatebox{90}{DS-V4-Flash}} & \multicolumn{3}{c}{\rotatebox{90}{Llama-3.1-8B}} & \multicolumn{3}{c}{\rotatebox{90}{L4-Maverick}} & \multicolumn{3}{c}{\rotatebox{90}{EXAONE-4.0}} & \multicolumn{3}{c}{\rotatebox{90}{Solar-100B}} & \multicolumn{3}{c}{\rotatebox{90}{Mi:dm-2.0}} & \multicolumn{3}{c}{\rotatebox{90}{EXAONE-3.5}} \\
\cmidrule(lr){3-5} \cmidrule(lr){6-8} \cmidrule(lr){9-11} \cmidrule(lr){12-14} \cmidrule(lr){15-17} \cmidrule(lr){18-20} \cmidrule(lr){21-23} \cmidrule(lr){24-26} \cmidrule(lr){27-29} \cmidrule(lr){30-32} \cmidrule(lr){33-35} \cmidrule(lr){36-38} \cmidrule(lr){39-41} \cmidrule(lr){42-44} \cmidrule(lr){45-47}
\textbf{Task} & & E & M & H & E & M & H & E & M & H & E & M & H & E & M & H & E & M & H & E & M & H & E & M & H & E & M & H & E & M & H & E & M & H & E & M & H & E & M & H & E & M & H & E & M & H \\
\midrule
\multirow{2}{*}{Array Formula} & EN & 83 & 67 & 60 & 80 & 72 & 58 & 80 & 60 & 46 & 73 & 47 & 28 & 72 & 52 & 29 & 92 & 83 & 57 & 64 & 47 & 20 & 58 & 36 & 20 & 66 & 47 & 22 & 10 & 3 & 2 & 54 & 36 & 15 & 19 & 6 & 5 & 18 & 1 & 3 & 3 & 1 & 3 & 1 & 0 & 0 \\
 & KO & 76 & 77 & 50 & 78 & 68 & 60 & 80 & 65 & 41 & 62 & 45 & 13 & 65 & 47 & 17 & 92 & 83 & 48 & 62 & 51 & 22 & 48 & 30 & 10 & 62 & 51 & 10 & 6 & 2 & 1 & 44 & 27 & 9 & 19 & 4 & 1 & 14 & 3 & 4 & 4 & 0 & 2 & 2 & 1 & 1 \\
\addlinespace[1.5pt]
\multirow{2}{*}{Causal DAG} & EN & 79 & 61 & 43 & 76 & 57 & 36 & 79 & 61 & 42 & 76 & 39 & 24 & 74 & 46 & 27 & 99 & 97 & 95 & 78 & 46 & 29 & 75 & 34 & 21 & 64 & 38 & 19 & 15 & 4 & 2 & 82 & 71 & 48 & 56 & 4 & 4 & 40 & 1 & 0 & 6 & 0 & 1 & 1 & 0 & 0 \\
 & KO & 78 & 62 & 41 & 75 & 54 & 38 & 79 & 61 & 34 & 68 & 41 & 18 & 72 & 41 & 23 & 100 & 98 & 93 & 78 & 48 & 24 & 71 & 32 & 19 & 64 & 34 & 16 & 8 & 2 & 0 & 78 & 69 & 57 & 46 & 9 & 4 & 36 & 3 & 2 & 4 & 0 & 0 & 1 & 0 & 0 \\
\addlinespace[1.5pt]
\multirow{2}{*}{Inequality} & EN & 98 & 92 & 80 & 100 & 85 & 70 & 92 & 88 & 63 & 55 & 40 & 23 & 87 & 72 & 51 & 82 & 62 & 34 & 78 & 61 & 29 & 67 & 60 & 28 & 73 & 61 & 35 & 1 & 0 & 0 & 7 & 2 & 0 & 67 & 25 & 17 & 42 & 25 & 7 & 0 & 0 & 0 & 0 & 0 & 0 \\
 & KO & 98 & 93 & 77 & 92 & 89 & 72 & 93 & 86 & 59 & 59 & 42 & 22 & 91 & 71 & 36 & 83 & 58 & 45 & 81 & 60 & 34 & 78 & 63 & 27 & 72 & 56 & 43 & 0 & 0 & 0 & 6 & 0 & 0 & 76 & 51 & 28 & 42 & 28 & 9 & 0 & 0 & 0 & 0 & 0 & 0 \\
\addlinespace[1.5pt]
\multirow{2}{*}{Minesweeper} & EN & 98 & 100 & 98 & 99 & 97 & 94 & 95 & 95 & 77 & 49 & 26 & 11 & 77 & 62 & 28 & 83 & 58 & 36 & 80 & 58 & 20 & 86 & 75 & 50 & 62 & 50 & 32 & 1 & 0 & 0 & 10 & 2 & 0 & 54 & 12 & 0 & 17 & 0 & 0 & 0 & 0 & 0 & 0 & 0 & 0 \\
 & KO & 96 & 95 & 94 & 99 & 96 & 92 & 91 & 94 & 73 & 51 & 27 & 9 & 81 & 61 & 18 & 84 & 59 & 32 & 74 & 43 & 19 & 86 & 74 & 51 & 58 & 26 & 22 & 1 & 0 & 0 & 8 & 2 & 0 & 46 & 7 & 1 & 32 & 1 & 0 & 0 & 0 & 0 & 0 & 0 & 0 \\
\addlinespace[1.5pt]
\multirow{2}{*}{Number Baseball} & EN & 100 & 99 & 98 & 99 & 95 & 97 & 97 & 74 & 62 & 62 & 40 & 22 & 81 & 58 & 39 & 87 & 65 & 44 & 86 & 59 & 42 & 83 & 66 & 46 & 60 & 53 & 29 & 2 & 1 & 0 & 15 & 5 & 1 & 43 & 12 & 8 & 24 & 5 & 2 & 0 & 0 & 0 & 0 & 0 & 0 \\
 & KO & 99 & 99 & 97 & 100 & 94 & 96 & 93 & 81 & 57 & 64 & 42 & 28 & 79 & 46 & 20 & 90 & 68 & 52 & 84 & 54 & 40 & 86 & 72 & 34 & 53 & 46 & 26 & 0 & 0 & 0 & 12 & 4 & 0 & 49 & 9 & 7 & 36 & 3 & 2 & 0 & 0 & 0 & 0 & 0 & 0 \\
\addlinespace[1.5pt]
\multirow{2}{*}{SAT Puzzle} & EN & 98 & 94 & 87 & 98 & 80 & 46 & 98 & 93 & 50 & 78 & 55 & 23 & 81 & 58 & 23 & 85 & 66 & 34 & 78 & 61 & 19 & 56 & 21 & 12 & 72 & 48 & 22 & 6 & 2 & 0 & 64 & 45 & 14 & 66 & 24 & 5 & 15 & 6 & 3 & 0 & 0 & 0 & 0 & 0 & 0 \\
 & KO & 92 & 91 & 80 & 95 & 66 & 28 & 98 & 86 & 33 & 71 & 63 & 17 & 73 & 53 & 27 & 91 & 72 & 36 & 92 & 48 & 20 & 53 & 25 & 8 & 59 & 38 & 15 & 4 & 1 & 0 & 57 & 44 & 17 & 60 & 13 & 2 & 11 & 1 & 1 & 0 & 0 & 0 & 1 & 0 & 0 \\
\addlinespace[1.5pt]
\multirow{2}{*}{Sudoku} & EN & 100 & 99 & 94 & 100 & 99 & 99 & 100 & 100 & 88 & 48 & 28 & 10 & 91 & 90 & 55 & 87 & 64 & 37 & 98 & 87 & 41 & 100 & 100 & 100 & 84 & 69 & 39 & 2 & 0 & 0 & 8 & 2 & 0 & 84 & 62 & 19 & 45 & 21 & 5 & 1 & 0 & 0 & 0 & 0 & 0 \\
 & KO & 100 & 100 & 100 & 100 & 100 & 100 & 100 & 100 & 96 & 49 & 31 & 8 & 94 & 87 & 48 & 87 & 68 & 32 & 98 & 87 & 53 & 100 & 99 & 100 & 80 & 64 & 44 & 1 & 0 & 0 & 6 & 2 & 0 & 92 & 73 & 15 & 68 & 30 & 14 & 0 & 0 & 0 & 1 & 0 & 0 \\
\addlinespace[1.5pt]
\multirow{2}{*}{Yacht Dice} & EN & 79 & 58 & 36 & 77 & 57 & 39 & 79 & 62 & 37 & 47 & 33 & 18 & 72 & 51 & 23 & 83 & 66 & 43 & 72 & 59 & 39 & 76 & 60 & 28 & 65 & 55 & 23 & 3 & 1 & 0 & 10 & 3 & 1 & 71 & 48 & 30 & 47 & 17 & 9 & 3 & 1 & 0 & 7 & 4 & 0 \\
 & KO & 76 & 60 & 45 & 75 & 54 & 37 & 79 & 54 & 35 & 48 & 31 & 22 & 68 & 43 & 22 & 84 & 63 & 49 & 71 & 56 & 30 & 76 & 55 & 26 & 69 & 38 & 23 & 3 & 0 & 0 & 8 & 3 & 0 & 60 & 32 & 21 & 49 & 21 & 8 & 11 & 1 & 1 & 2 & 1 & 0 \\
\addlinespace[1.5pt]
\midrule
\multirow{2}{*}{Cipher} & EN & 95 & 96 & 95 & 98 & 96 & 85 & 91 & 87 & 76 & 71 & 10 & 2 & 45 & 39 & 22 & 92 & 66 & 37 & 92 & 78 & 29 & 93 & 53 & 21 & 88 & 91 & 44 & 2 & 0 & 0 & 2 & 0 & 0 & 46 & 44 & 17 & 34 & 25 & 5 & 0 & 0 & 0 & 0 & 0 & 0 \\
 & KO & 97 & 94 & 90 & 87 & 68 & 53 & 95 & 54 & 40 & 17 & 0 & 0 & 0 & 0 & 0 & 55 & 18 & 7 & 21 & 5 & 0 & 2 & 0 & 0 & 15 & 1 & 1 & 0 & 0 & 0 & 9 & 4 & 0 & 0 & 0 & 0 & 0 & 0 & 0 & 0 & 0 & 0 & 0 & 0 & 0 \\
\addlinespace[1.5pt]
\multirow{2}{*}{Cryptarithmetic} & EN & 99 & 95 & 94 & 97 & 91 & 81 & 94 & 89 & 63 & 55 & 28 & 13 & 63 & 41 & 14 & 85 & 61 & 33 & 74 & 34 & 18 & 71 & 49 & 25 & 64 & 57 & 37 & 3 & 1 & 0 & 12 & 4 & 1 & 59 & 30 & 11 & 9 & 1 & 0 & 0 & 0 & 0 & 0 & 0 & 0 \\
 & KO & 99 & 95 & 98 & 99 & 94 & 92 & 99 & 91 & 74 & 42 & 20 & 7 & 16 & 4 & 0 & 92 & 64 & 39 & 75 & 35 & 19 & 20 & 17 & 9 & 68 & 58 & 42 & 4 & 0 & 0 & 4 & 0 & 0 & 0 & 0 & 0 & 5 & 2 & 0 & 0 & 0 & 0 & 0 & 0 & 0 \\
\addlinespace[1.5pt]
\midrule
Kinship & KO & 68 & 36 & 33 & 59 & 54 & 77 & 80 & 68 & 59 & 14 & 8 & 5 & 17 & 22 & 17 & 44 & 26 & 19 & 57 & 26 & 21 & 7 & 7 & 7 & 24 & 20 & 17 & 4 & 1 & 0 & 10 & 6 & 4 & 9 & 6 & 4 & 5 & 3 & 5 & 5 & 2 & 4 & 4 & 3 & 0 \\
\addlinespace[1.5pt]
Saju & KO & 99 & 89 & 83 & 94 & 82 & 56 & 90 & 74 & 71 & 44 & 19 & 8 & 62 & 44 & 15 & 75 & 47 & 24 & 70 & 45 & 34 & 17 & 17 & 19 & 73 & 48 & 28 & 2 & 1 & 0 & 7 & 3 & 0 & 1 & 3 & 1 & 9 & 5 & 5 & 3 & 3 & 2 & 1 & 0 & 0 \\
\addlinespace[1.5pt]
Time & KO & 98 & 94 & 90 & 93 & 76 & 69 & 97 & 94 & 93 & 58 & 18 & 7 & 67 & 37 & 13 & 84 & 51 & 34 & 80 & 51 & 29 & 78 & 61 & 40 & 69 & 41 & 11 & 8 & 2 & 0 & 35 & 8 & 4 & 67 & 32 & 9 & 68 & 36 & 8 & 65 & 15 & 8 & 30 & 7 & 1 \\
\addlinespace[1.5pt]
Jamo Composition & KO & 95 & 88 & 79 & 67 & 37 & 37 & 79 & 56 & 37 & 26 & 8 & 3 & 3 & 1 & 0 & 72 & 45 & 22 & 32 & 7 & 2 & 18 & 6 & 2 & 41 & 10 & 1 & 2 & 1 & 0 & 5 & 1 & 0 & 0 & 0 & 0 & 9 & 2 & 0 & 1 & 1 & 0 & 0 & 0 & 0 \\
\addlinespace[1.5pt]
Korean Units & KO & 99 & 97 & 96 & 95 & 86 & 85 & 47 & 22 & 16 & 38 & 20 & 7 & 52 & 28 & 23 & 82 & 61 & 47 & 9 & 0 & 0 & 92 & 77 & 85 & 84 & 70 & 72 & 0 & 0 & 0 & 8 & 3 & 0 & 48 & 17 & 16 & 45 & 1 & 1 & 0 & 0 & 0 & 0 & 0 & 0 \\
\addlinespace[1.5pt]
\midrule
\multicolumn{2}{@{}l}{\textbf{Macro avg}} & 92.0 & 85.2 & 77.5 & 89.3 & 77.9 & 67.9 & 88.2 & 75.8 & 56.9 & 53.0 & 30.4 & 13.9 & 63.3 & 46.2 & 23.6 & 83.6 & 62.8 & 41.2 & 71.4 & 48.2 & 25.3 & 63.9 & 47.6 & 31.5 & 63.6 & 46.8 & 26.9 & 3.5 & 0.9 & 0.2 & 22.4 & 13.8 & 6.8 & 45.5 & 20.9 & 9.0 & 28.8 & 9.6 & 3.7 & 4.2 & 1.0 & 0.8 & 2.0 & 0.6 & 0.1 \\
\addlinespace[1.5pt]
\bottomrule
\end{tabular}
\caption{Accuracy (\%) for all 25 tasks $\times$ 3 tiers and all 15 models (E/M/H = Easy/Medium/Hard); model order matches Table~\ref{tab:main-results}. Bilingual tasks show English and Korean on adjacent rows; horizontal rules separate direct translations, script adaptations, and Korean-only tasks. Macro avg is the per-task macro average over all 25 task--languages. Opus 4.8 shows a tier inversion on Kinship (59/54/77), one of the 45 model--task pairs \protect\jydel{non-monotone}\jy{without a strict Easy$>$Medium$>$Hard decrease} noted in Figure~\ref{fig:tiers}.}
\label{tab:full-results}
\end{sidewaystable*}